\documentclass[preprint,authoryear,a4paper]{elsarticle}

\usepackage{verbatim}

\usepackage{geometry}

\usepackage{color,soul}

\usepackage[hidelinks]{hyperref}
\hypersetup{
  colorlinks   = true, 
  urlcolor     = blue,
  linkcolor    = blue, 
  citecolor    = blue 
}

\usepackage{graphicx}
\graphicspath{{./images/}}
\DeclareGraphicsExtensions{.pdf,.png,.jpg}
\usepackage{placeins}
\usepackage{caption}
\usepackage{subcaption}

\usepackage{booktabs}
\usepackage[table,xcdraw]{xcolor}
\usepackage{graphicx}
\usepackage{adjustbox}
\usepackage{multirow}

\usepackage{amssymb}
\usepackage{gensymb}
\usepackage[mathscr]{euscript}
\usepackage{bm}
\usepackage{mathtools}
\usepackage{nccmath}
\usepackage{amsmath}
\usepackage{braket}

\usepackage{algorithm}
\usepackage{algpseudocode}

\usepackage{lineno, hyperref}

\usepackage{lipsum}

\newcommand{\red}[1]{#1}

\journal{ISPRS Journal of Photogrammetry and Remote Sensing}

\begin{document}

\begin{frontmatter}

    \title{Structure-aware Completion of Photogrammetric Meshes in Urban Road Environment}

    \author[swjtu]{Qing Zhu}
    \author[swjtu]{Qisen Shang}
    \author[swjtu]{Han Hu\corref{cor1}}
    \author[swjtu]{Haojia Yu}
    \author[bnu]{Ruofei Zhong}
    \cortext[cor1]{Corresponding Author: han.hu@swjtu.edu.cn}

    \address[swjtu]{Faculty of Geosciences and Environmental Engineering, Southwest Jiaotong University, Chengdu, China}
    \address[bnu]{Beijing Advanced Innovation Center for Imaging Technology, Capital Normal University, Beijing, China}
    \begin{abstract}     
        Photogrammetric mesh models obtained from aerial oblique images have been widely used for urban reconstruction.
        However, photogrammetric meshes suffer from severe texture problems, particularly in typical road areas, owing to occlusion.
        This paper proposes a structure-aware completion approach to improve mesh quality by seamlessly removing undesired vehicles.
        Specifically, a discontinuous texture atlas is first integrated into a continuous screen space by rendering trough a graphics pipeline. The rendering also records the necessary mapping for deintegration to the original texture atlas after editing.
        Vehicle regions are masked by a standard object detection approach, namely, Faster RCNN.
        Subsequently, the masked regions are completed, guided by the linear structures and regularities in the road region; this is implemented based on PatchMatch.
        Finally, the completed rendered image is deintegrated to the original texture atlas, and the triangles for the vehicles are also flattened so that improved meshes can be obtained.
        Experimental evaluation and analysis are conducted on three datasets, which were captured with different sensors and ground sample distances.
        The results demonstrate that the proposed method can produce quite realistic meshes after removing the vehicles.
        The structure-aware completion approach for road regions outperforms popular image completion methods, and an ablation study further confirms the effectiveness of the linear guidance.
        It should be noted that the proposed method can also handle tiled mesh models for large-scale scenes.
        Code and datasets are available at the project website\footnote{\url{https://vrlab.org.cn/~hanhu/projects/mesh}}.
    \end{abstract}

\end{frontmatter}


\section{Introduction}
\label{s:intro}

The demand for automatic modeling of city-scale urban environments has recently attracted increasing interest  \citep{remondino2017critical}. Specifically, high-quality 3D reconstruction of road networks, which form the skeleton of urban scenes, is useful in a variety of applications, including navigation maps, autonomous driving, and urban planning \citep{yang2017computing,chen2019higher,wenzel2019simultaneous}. Recently, massive airborne datasets have been collected for several cities around the world \citep{google2020earth,sugarbaker20143d,isenburg2020open}. Although airborne light detection and ranging (LiDAR) has been widely used in the last two decades \citep{kada20093d,haala2010update}, advances in large-scale structure-from-motion (SFM)  \citep{schonberger2016structure} and multi-view stereo (MVS) \citep{vu2009towards,jancosek2011multi} pipelines have enabled the automatic generation of city-scale triangular surface models from aerial oblique images, which are also enriched with high-resolution textures.  

Aerial oblique images are arguably the most widely used datasets for 3D modeling of urban environments \citep{google2020earth}; however, photogrammetric point clouds and meshes are geometrically less accurate and regular than LiDAR points at similar resolutions \citep{nex2014photogrammetric, hu2016texture}. In addition, photogrammetric meshes are also commonly contaminated by occluded objects and topological defects \citep{verdie2015lod, hu2016stable}, particularly in road areas. These issues require a practical approach to improve the quality of existing photo-realistic meshes from aerial oblique images. In particular, the major objective of this study is to remove vehicles from both the geometries and textures of mesh models so that clean results can be obtained for further applications. Despite the recent progress in the processing of textures of mesh models \citep{prada2018gradient}, MVS pipelines for photogrammetric meshes of road regions cause certain problems that should be addressed.

\begin{figure}[htb]
	\centering
	\subcaptionbox{Defects of meshes.}[0.49\linewidth]{\includegraphics[width=\linewidth]{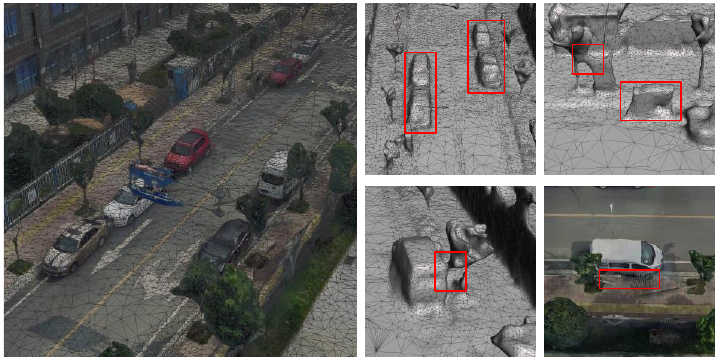}}
	\subcaptionbox{Discontinuous UV atlases}[0.49\linewidth]{\includegraphics[width=\linewidth]{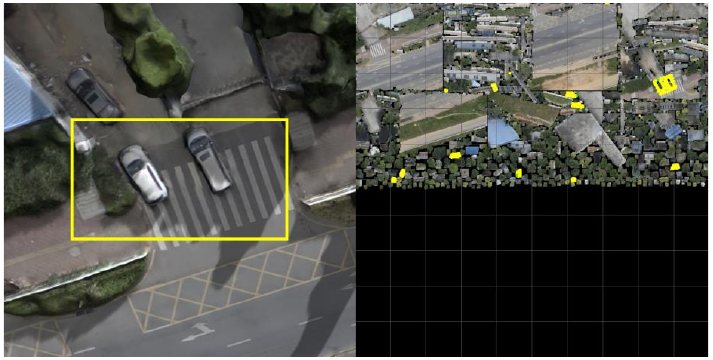}}
	\caption{Common problems of photogrammetric meshes.}
	\label{fig:mesh_problems}
\end{figure}

\textit{1) Occlusion and dynamic objects.} Although penta-view aerial oblique camera systems can capture ground objects from multiple viewpoints \citep{remondino2015oblique}, occlusions on the ground, particularly  in road areas, are inevitable in urban environments with dense building rise-ups. The geometries of photogrammetric meshes are generally noise-laden, and the textures are severely blurred (Figure \ref{fig:mesh_problems}a). In addition, existing MVS pipelines cannot handle dynamic objects, \textit{ for example } vehicles. We argue that it is hardly possible to resolve these issues in an MVS pipeline for aerial oblique images because the aforementioned defects are inherently embedded in the images. Therefore, direct editing of the meshes to improve quality may be the most practical solution.

\textit{2) Discrete and discontinuous atlases for textures.} A photogrammetric surface mesh, as a 2D manifold embedded in 3D space, should be unwrapped to 2D planar space so that it can be consumed by the textures. This step is also termed surface parametrization \citep{levy2002least}. Although the existence of a continuous conformal map for any 2D manifold has been proved, in practice, photogrammetric meshes should be separated into different segments (\textit{e.g., } charts) and packed into one or multiple texture images (\textit{e.g., } atlases) (Figure \ref{fig:mesh_problems}b). Segmentation is generally used to alleviate distortion in surface parametrization for large patches and to achieve fitting into a single image with a limited field of view \citep{waechter2014let}. It is quite difficult to directly process texture images, which are discrete and discontinuous.  

To resolve these issues, we propose a structure-aware completion method for the textures of photogrammetric meshes to improve the mesh quality of urban roads.
The intuitive principle is to detect vehicles, directly flatten the geometries, and replace the textures according to repeated patterns of urban roads.
Specifically, we first transform the discrete texture atlas to a continuous screen space through the graphics pipeline.
Subsequently, vehicles are detected and masked as voids using publicly available deep learning approaches \cite{ren2015faster}.
Then, the void areas are automatically completed using the proposed structure-aware image completion method.
Finally, the completed textures are remapped to the original texture atlases, and the triangles corresponding to the voids are flattened.

In summary, the major advantages of the proposed mesh correction approaches are the following: 1) \red{a practical solution for the editing of photogrammetric mesh models with discontinuous texture atlases by directly rendering through graphic pipelines}, and 2) structure-aware image completion methods to remove vehicles on the roads of urban environments.
The rest of this paper is organized as follows: Section \ref{s:related_work} provides a brief review of related work. Section \ref{s:method_overview} introduces the mesh completion workflow. Sections \ref{s:texture_editing}, \ref{s:detection}, and \ref{s:image_completion} elaborate the details of proposed method. Experimental evaluations are presented in Section \ref{s:results}. Finally, the last section concludes the paper.

\section{Related work}
\label{s:related_work}

The most relevant literature includes 1) texture mapping and processing \citep{waechter2014let,prada2018gradient}, 2) object detection \citep{ren2015faster,redmon2016you}, and 3) image completion \citep{he2012statistics}.

\paragraph{1) Texture mapping and processing} 
Currently, massive collections of city-scale aerial oblique images \citep{remondino2015oblique} have been obtained.
With the advances in bundle adjustment \citep{hu2015reliable,verykokou2018oblique} and dense image matching \citep{hirschmuller2008stereo,hu2016texture} for aerial oblique images, high-density photogrammetric point clouds can be obtained, from which surface meshes can be constructed \citep{kazhdan2013screened,jancosek2011multi}.
These meshes are then enriched with high-resolution textures by mapping the color information from the original images \citep{zhou2014color,bi2017patch,waechter2014let}.
In theory, any manifold surface immersed in 3D space can be continuously unwrapped in 2D space \citep{crane2013digital}; this process is termed surface parametrization \citep{levy2002least}.
Unfortunately, however, a continuous parametrization inevitably leads to significantly distorted shapes in 2D space \citep{sorkine2002bounded}, particularly for large and irregular meshes.
Therefore, the texture mapping of a photogrammetric mesh is intentionally segregated into different segments (\textit{charts}), and different parts are then packed into a single or several texture images (\textit{atlases}) \citep{levy2002least,liu2019atlas,limper2018box}.
The boundaries for different charts are commonly known as seam lines.
Several similar strategies are available to generate seam lines or, equivalently, charts, including variational shape approximation \citep{cohen2004variational}, local geometries \citep{allene2008seamless,zhang2020robust}, and multi-view geometric and photometric consistencies \citep{waechter2014let,bi2017patch}.

The processing of discontinuous texture atlases is quite difficult \citep{yuksel2019rethinking}.
To alleviate color differences in the continuous image space (\textit{i.e., } image mosaicking), we could either estimate the color transfer function based on the overlapping region \citep{yu2017auto,hu2019color}, or reformulate the problem as a gradient-domain blending problem \citep{agarwala2007efficient,kazhdan2010distributed}.
However, in discontinuous texture atlases, it is difficult to detect or even define the overlaps and gradient breaks on the seam lines.
Thus,  \cite{waechter2014let} only locally blends the color in a region with a fixed width along seam lines.
To resolve this discontinuity, \cite{prada2018gradient} and \cite{liu2017seamless} formulate the problem in a meticulously selected continuous function space based on the finite element approach.
However, these approaches cannot handle varying mesh typologies and large texture sizes (e.g., $8192\times8192$) \citep{waechter2014let}, and therefore, they are probably not suitable for editing photogrammetric meshes.
Instead, we propose a practical approach to edit mesh textures by efficiently generating a continuous mapping through orthogonal rendering of the meshes.

\paragraph{2) Object detection}
Traditional object detection methods generally use low-level features, such as Harr-like features \citep{papageorgiou1998general}, local binary patterns \citep{ojala2002multiresolution}, and histograms of oriented gradients \citep{dalal2005histograms}.
Some notable approaches include the V-J detector \citep{viola2004robust} and deformable part models \citep{felzenszwalb2009object}.
However, owing to the semantic gap between low-level vision features and high-level semantic information, only low performance has been achieved for a considerable time \citep{girshick2014rich}.
With the re-invention of deep neural networks, particularly deep convolutional neural networks, learned features \citep{simonyan2014very} pre-trained on large datasets have demonstrated superior performance compared with previously known shallow methods.
A typical approach, known as the regional convolutional neural network (RCNN) \citep{girshick2014rich}, can perform unsupervised detection  of te bounding boxes of certain salient objects  \citep{uijlings2013selective}, aggregate the features in the bounding box through regional pooling, and classify each bounding box into specified categories.
This strategy is further improved by reusing the feature maps \citep{girshick2015fast} and learning the generation of bounding boxes \citep{ren2015faster}; this strategy is known as Faster RCNN.

These approaches consist of two separate stages: detection and classification of bounding boxes.
To improve efficiency, a more concise one-stage approach is proposed \citep{redmon2016you}, namely "you only look once"
Its principle is to tessellate an image into regular grids, and regress each grid to the corresponding locations and most likely classes.
Similar approaches have been proposed to further improve effectiveness \citep{liu2016ssd, redmon2017yolo9000, redmon2018yolov3}.
Owing to the superior performance of deep learning in object detection, we also use a standard approach \citep{ren2015faster} to detect vehicles on roads.

\paragraph{3) Image completion}
With regard to filling void image regions, extensive research has been conducted on image inpainting and completion. The corresponding methods are based either on partial differential equations (PDEs) or on sampling. 

\cite{bertalmio2000image} repaired images by diffusing the edge information of the region of interest (ROI) to unknown regions.
A similar approach was proposed using total variation \citep{shen2002mathematical}, and was improved through an anti-aliasing strategy \citep{aubert2006mathematical}.
However, PDE-based methods cannot fill large voids, as they are only driven by local information. 

Regarding sample-based approaches, \cite{criminisi2004region} selected the best matching patch according to the isophote. This method was improved by considering the weighted averaging of multiple patches \citep{wong2008nonlocal}.
A milestone on sample-based image completion  is patch matching \citep{barnes2009patchmatch}, which significantly accelerates the search for the most similar patches through random guess and expansion.
Several state-of-the-art methods are based on patch matching: \cite{he2012statistics} extracted transnational regularities using  the statistics of patch offsets, and \cite{huang2014image} further considered affine deformations.
Although sample-based methods can repair large missing images, they are  limited to structured linear patterns \citep{iizuka2017globally}, which are probably the most common patterns for urban roads.

Recently, image completion based on deep learning \citep{graves2013generating, yeh2017semantic, iizuka2017globally} has made progress by using generative models \citep{iizuka2017globally, yu2018generative, nazeri2019edgeconnect}. However, deep learning approaches rely on massive training data, which are difficult to obtain; otherwise, it is difficult to synthesize high-resolution images \citep{wu2017survey}. In this paper, we propose a structure-aware image completion that directly uses linear features to guide the search for similar patches.

\section{Structure-aware completion of photogrammetric meshes for urban roads}
\label{approach}
\subsection{Overview and problem setup}
\label{s:method_overview}

\subsubsection{Overview of the approach}

\begin{figure}[H]
    \centering
    \includegraphics[width=\textwidth]{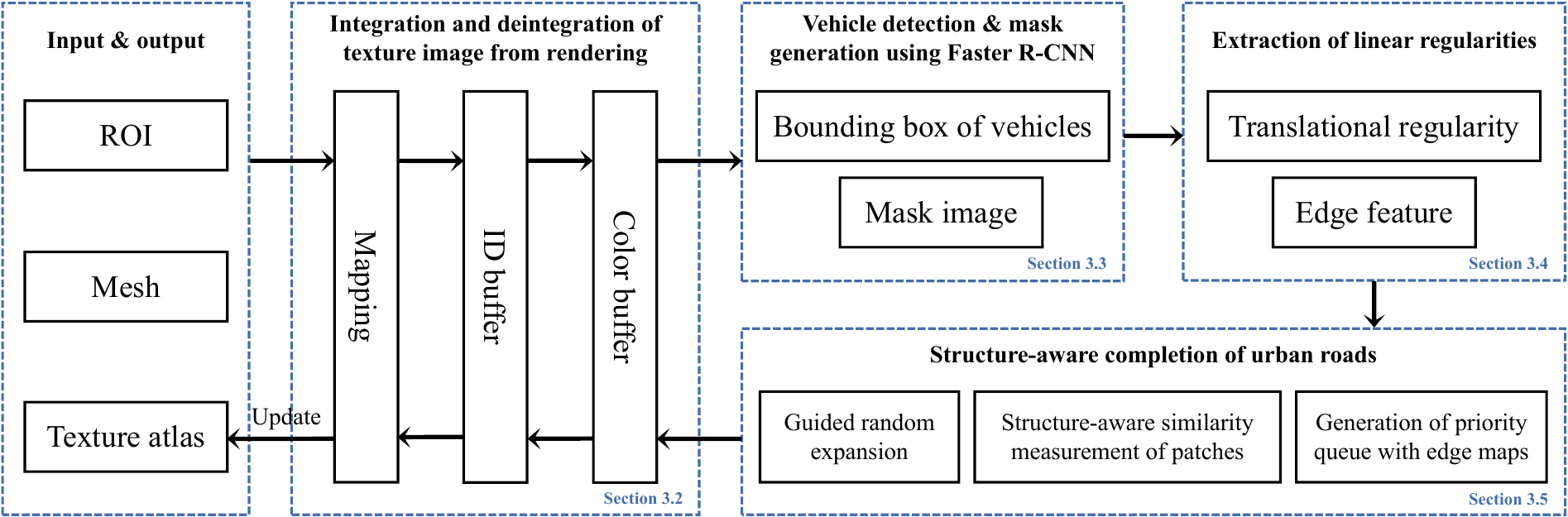}
    \caption{Workflow of photogrammetric mesh correction in road areas. First, the discontinuous texture of the selected ROI is integrated by rendering, vehicles are detected, and a mask image is generated using Faster R-CNN. Then, regularities are extracted and used to guide and constrain the image completion process. Subsequently, the corrected integrated image is deintegrated by rendering to obtain the corrected texture atlas. Finally, the texture atlas is replaced with a corrected texture atlas to achieve texture correction of the mesh.}
    \label{fig:flow_of_method}
\end{figure}

To overcome the problem caused by discrete and discontinuous textures, we establish a one-to-one mapping between the atlas and the orthogonally rendered image, whereby vehicle regions are extracted and filled.
In addition, we apply a linear feature, which is the most common feature in urban roads, to guide and constrain image completion and thus improve the output linear structures.
The overall workflow of the proposed method is shown in Figure \ref{fig:flow_of_method}; it consists of five parts.
Beginning with the textured meshes and an input ROI, we first render the geometry primitives (e.g., triangles) in the ROI to two buffers: an ID buffer that records the primitive IDs, and a color buffer that records the color information. Subsequently, we establish a mapping between a texel (a pixel in the texture atlas) and a pixel of color buffer using the corresponding primitive IDs.
After texture integration, we apply Faster R-CNN to the integrated image to detect vehicles and generate a mask according to the bounding boxes of the detected objects. Subsequently, we complete the image with a mask using the proposed algorithm, which involves two sequential steps: detecting the road direction from translational regularities by RANSAC, and extracting edges from the road image. The proposed algorithm uses this information to guide and constrain the completion process. Finally, the completion result is used, and the pipeline is rendered to update the texture atlas so that automatic correction of the textured model can be achieved.

\subsubsection{Problem setup}

\begin{figure}[htb]
	\centering
	\subcaptionbox{2-Manifold geometry mesh $\mathcal{M}$}[0.32\linewidth]{\includegraphics[width=\linewidth]{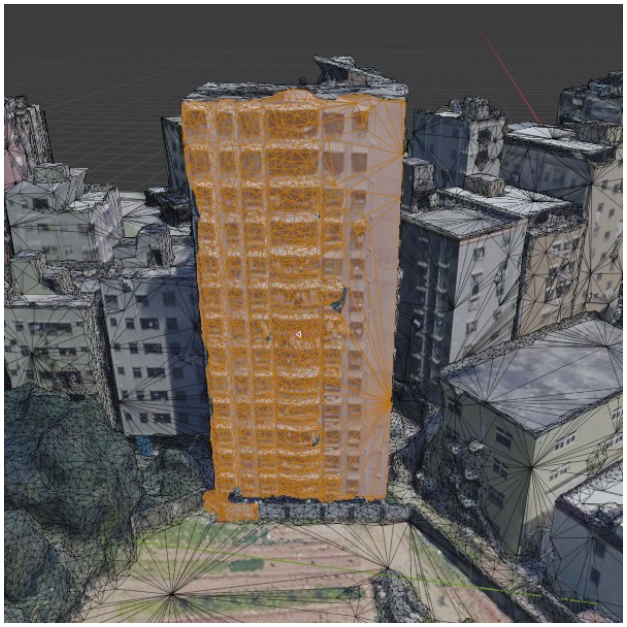}}
	\subcaptionbox{2D UV mesh $\mathcal{M}'$}[0.32\linewidth]{\includegraphics[width=\linewidth]{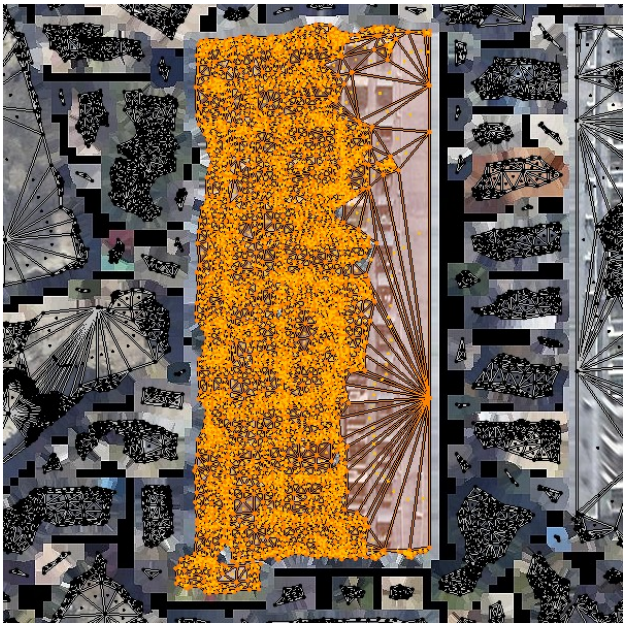}}
	\subcaptionbox{Texture image $\mathcal{I}$}[0.32\linewidth]{\includegraphics[width=\linewidth]{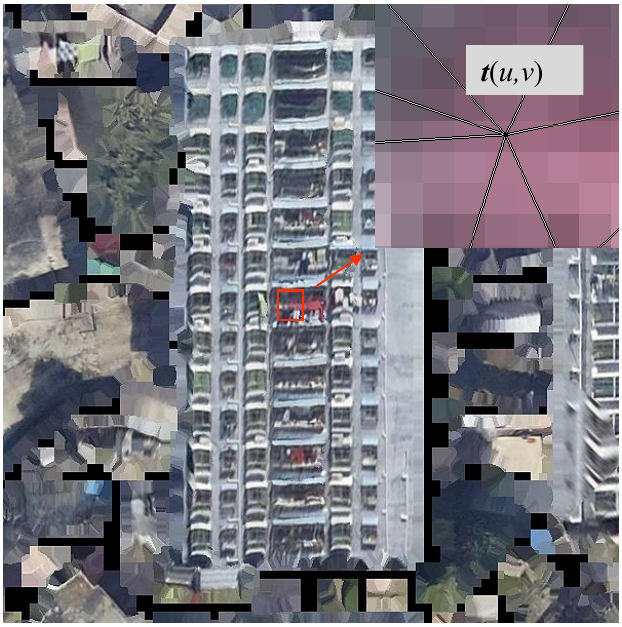}}
	\caption{A textured mesh in this paper consists of three parts: a geometry mesh $\mathcal{M}$, an UV mesh $\mathcal{M}'$, and a texture image $\mathcal{I}$.}
	\label{fig:mesh}
\end{figure}

More formally, the inputs consist of a 2-manifold geometry mesh  $ \mathcal{M}(V,F) $ and a 2D UV mesh $ \mathcal{M}'(V',F') $ \citep{prada2018gradient}, as shown in Figures \ref{fig:mesh} a and b.
$ V\in\mathbb{R}^{3 \times N} $ and $V'\in\mathbb{R}^{2 \times N'}$ are the vertices of the meshes.
In general, the number of vertices is not equal to $N \ne N'$ because of the seam lines in the mesh $ M $.
$F\in\mathbb{Z}^{3\times M}$ and $F'\in\mathbb{Z}^{3\times M'}$ are the facets; each column of $F$ and $F'$ records three indices into vertices $V$ and $V'$, respectively.
Although the indices may be different, the order and number of the facets recorded in $F$ and $F'$ are the same, that is, $M=M'$.
There is also a texture image $\mathcal{I}$ associated with the UV mesh $\mathcal{M}'$. To avoid confusion, we term the coordinates on the texture image \textit{texel} $\boldsymbol{t}(u,v)$ (Figure \ref{fig:mesh}c) \citep{prada2018gradient} rather than \textit{pixel} $\boldsymbol{p}(x,y)$ for normal images.
The purpose of this study is to flatten the geometries by modifying the vertices $V$ of the mesh $\mathcal{M}$, and to correct the texture image $\mathcal{I}$ so that cleaner street scenes can be obtained in an urban environment.

\subsection{Integration and deintegration of texture image}
\label{s:texture_editing}

\begin{figure}[h]
	\centering
	\includegraphics[width=\linewidth]{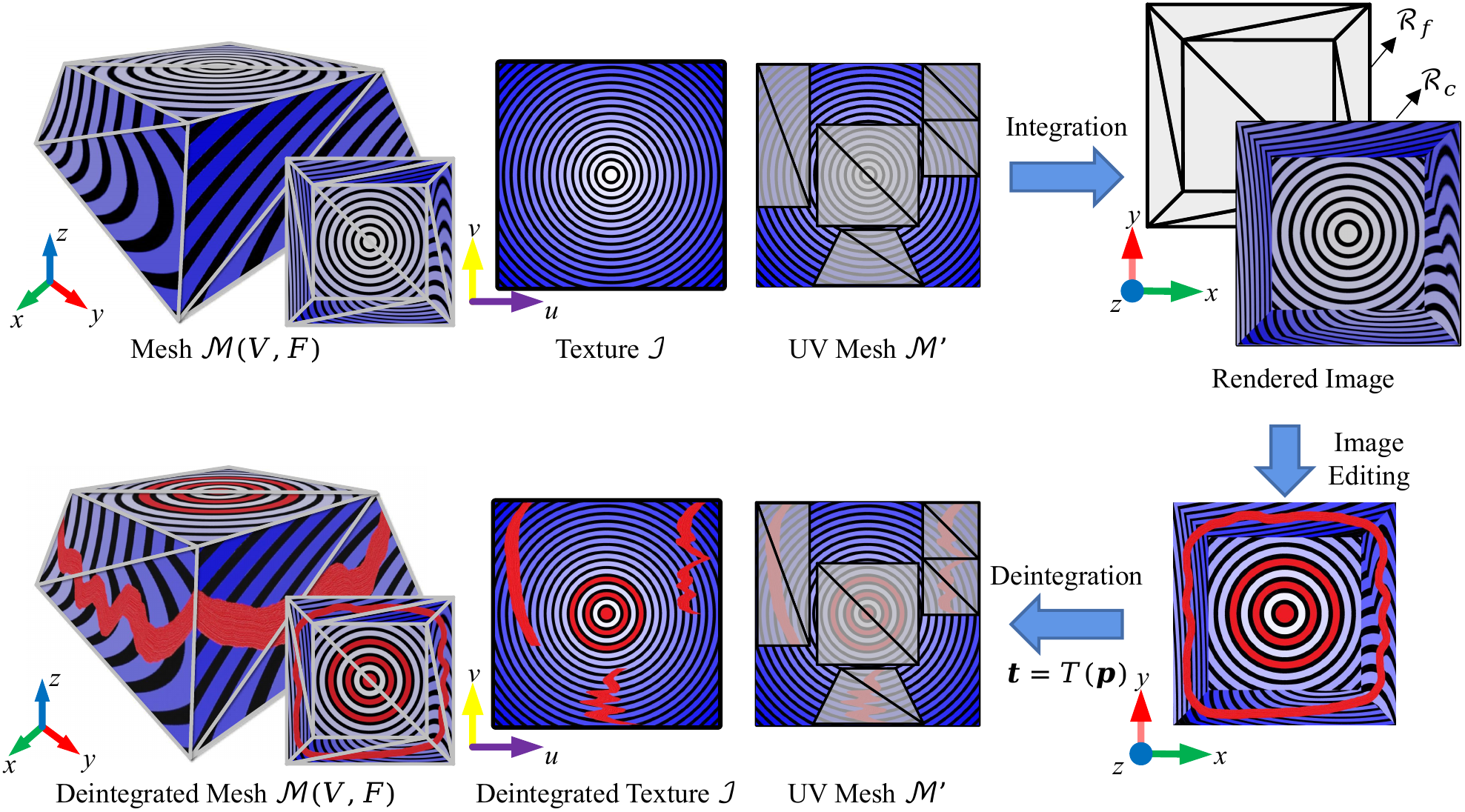}
	\caption{Integration and deintegration of texture image. Beginning with a mesh $\mathcal{M}$, a UV mesh $\mathcal{M}'$, and the texture $\mathcal{I}$, we first render the mesh in the ROI to two rasters: the color $\mathcal{R}_c$ and the facet raster $\mathcal{R}_f$. $R_c$ and $R_f$ record the grayscale values and facet numbers from the original meshes, respectively. After editing $\mathcal{R}_c$, we estimate a mapping $\boldsymbol{t}=T(\boldsymbol{p})$ using $\mathcal{R}_f$ and deintegrating the edited pixels to the texture image.}
	\label{fig:tex_integration}
\end{figure}

As discontinuous texture images are difficult to process, we first integrate the texels in a specified ROI into a continuous image, by directly rendering the textured mesh models \citep{zhu2020leveraging}.
After performing the correction steps in the continuous image space, we deintegrate the modified pixels to the corresponding texels using the methods described below.
Figure \ref{fig:tex_integration} shows the integration and deintegration processes of the texture image.

\subsubsection{Texture integration}
\label{subs:texture_integration}

The graphics render pipeline can efficiently project the mesh models to the screen space, and shade the fragment on the screen from the texture.
Despite the textured mesh models, two matrices are required that define the projection from the model space to screen viewport: the projection matrix $\mathbf{P}$ and the view matrix $\mathbf{V}$. As in \citep{zhu2020leveraging}, we use the \textit{ortho} and \textit{perspective} routines in GLM \citep{glm2019opengl} for the projection $\mathbf{P}$ and view $\mathbf{V}$ matrix, respectively.
The parameters of the \textit{ortho} and \textit{perspective} routines can be intuitively determined through the bounding box of the mesh models in the selected ROI.
In addition, the geometries outside the ROI are discarded in the following processing.

The direct output of the render pipeline is a color raster $\mathcal{R}_c$, which samples the texture by bilinear interpolation of the texels.
However, if only $\mathcal{R}_c$ is available, the information required to map between $\mathcal{R}_c$ and the original texture image $\mathcal{I}$ is insufficient.
Therefore, we also allocate another raster $\mathcal{R}_f$ in the render pipeline.
Each pixel $\boldsymbol{p}(x,y) \in \mathbb{Z}$ in $\mathcal{R}_f$ stores the facet index $f$ of the mesh  $F\in\mathcal{M}$ or, equivalently, the UV mesh $F'\in\mathcal{M}'$.
It should be noted that $\mathcal{R}_f$ and the color raster $\mathcal{R}_c$ are obtained simultaneously in a single rendering frame.
The \textit{gl\_PrimitiveID} in the fragment language of OpenGL is a constant that contains the index of the current primitive in the rendering pipeline, that is, $f=gl\_PrimitiveID$.
Similar inputs are also available for other graphic application programming interfaces.
Therefore, both $\mathcal{R}_c$ and $\mathcal{R}_f$, as shown in Figure \ref{fig:tex_integration}, can be obtained in real time.

\subsubsection{Texture deintegration}
\label{subs:model_update}

To map the edited image $\mathcal{R}_c$ back to the texture image $\mathcal{I}$, a one-to-one mapping should be \red{established} between the two images, that is, $\boldsymbol{t}=T(\boldsymbol{p})$, as shown in Figure \ref{fig:tex_integration}. Although the transformation from mesh $\mathcal{M}$ to UV mesh $\mathcal{M}'$ is generally assumed to be conformal \citep{levy2002least}, we relax this assumption to an affine transformation because of numerical rounding or other factors.
Therefore, we can use the trilinear coordinates $\boldsymbol{c}(a,b,c)$ \citep{wolfram2020trilinear} to interpolate $\eta(\cdot)$ the position inside the same facet $f$ for the mesh $\mathcal{M}$, UV mesh $\mathcal{M}'$, and rendered image $\mathcal{R}_c$.

\begin{figure}[H]
	\centering
	\includegraphics[width=\linewidth]{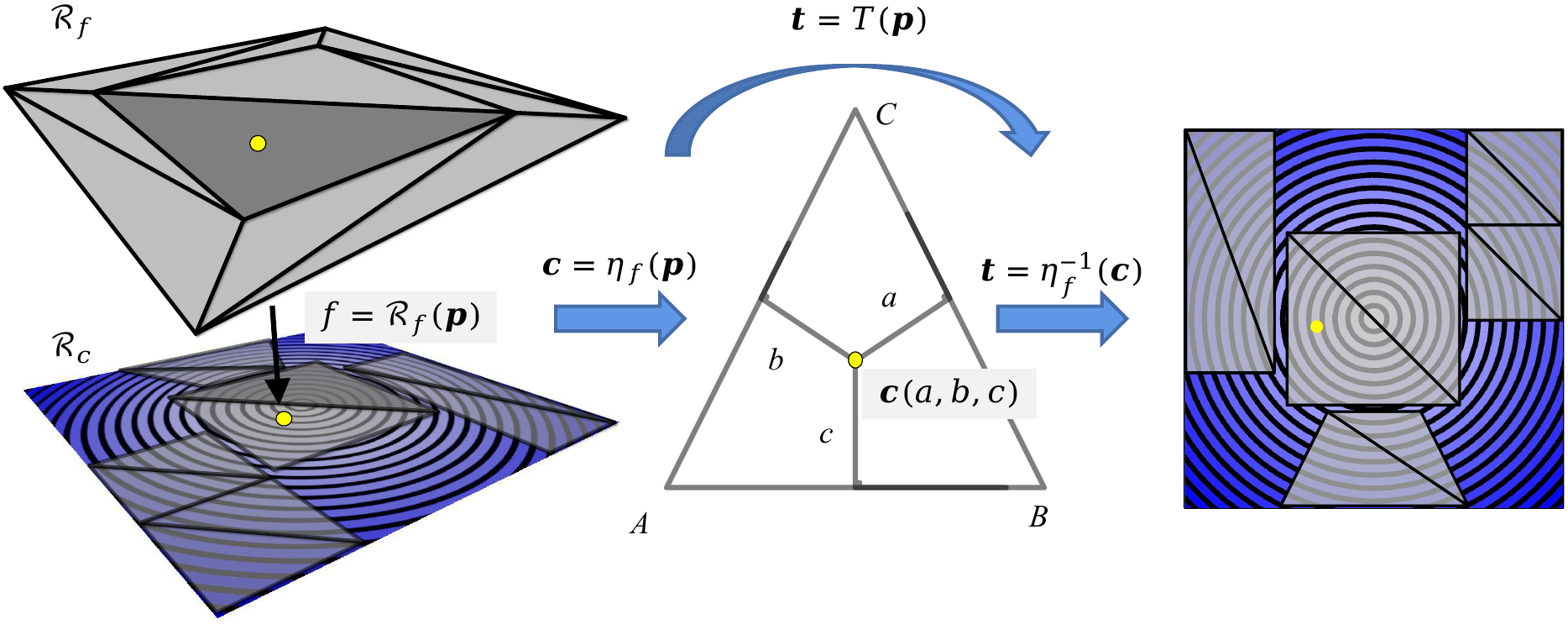}
	\caption{Mapping between rendered image $\mathcal{R}_c$ and texture image $\mathcal{I}$ using trilinear coordinates.}
	\label{fig:trilinear}
\end{figure}

Figure \ref{fig:trilinear} shows the process to establish the transformation $\boldsymbol{t}=T(\boldsymbol{p})$.
For each modified pixel $\boldsymbol{o}\in\mathcal{R}_c$, the corresponding facet index is directly loaded from $\mathcal{R}_f$ as $f=\mathcal{R}_f(\boldsymbol{p})$.
The trilinear coordinates of pixel $\boldsymbol{p}$ inside the corresponding triangle $f$ can be computed in closed form as $\boldsymbol{c}=\eta_f(\boldsymbol{p})$ \citep{wolfram2020trilinear}.
Then, in the same facet of the UV mesh $\mathcal{M}'$, the inverse mapping directly yields the position of the texel as $\boldsymbol{t}=\eta_f^{-1}(\boldsymbol{c})$.
In fact, inside each facet, the transformation $\boldsymbol{t}=T(\boldsymbol{p})$ is equivalent to an affine transformation.
In the implementation, the mapping $T(\cdot)$ is pre-computed for each triangle and stored pixel-wise.

Another practical issue for the integration and deintegration processes is related to the data structure of the mesh models.
The mesh models are tiled into small fragments and organized in a tree structure for different LODs.
Each small tile is associated with a texture.
In the implementation, the UV Mesh $\mathcal{M}'$ also contains an array to store the atlas index.
Moreover, multiple transformations $T(\cdot)$ are used to account for multiple texture atlases.
In fact, even without tiling, a single mesh can also contain multiple texture images; the same strategy is used to handle the above issue.

\subsection{Vehicle detection and mask generation using Faster R-CNN}
\label{s:detection}

\begin{figure}[h]
	\centering
	\subcaptionbox{Rendered image $\mathcal{R}_c$}[0.32\linewidth]{\includegraphics[width=\linewidth]{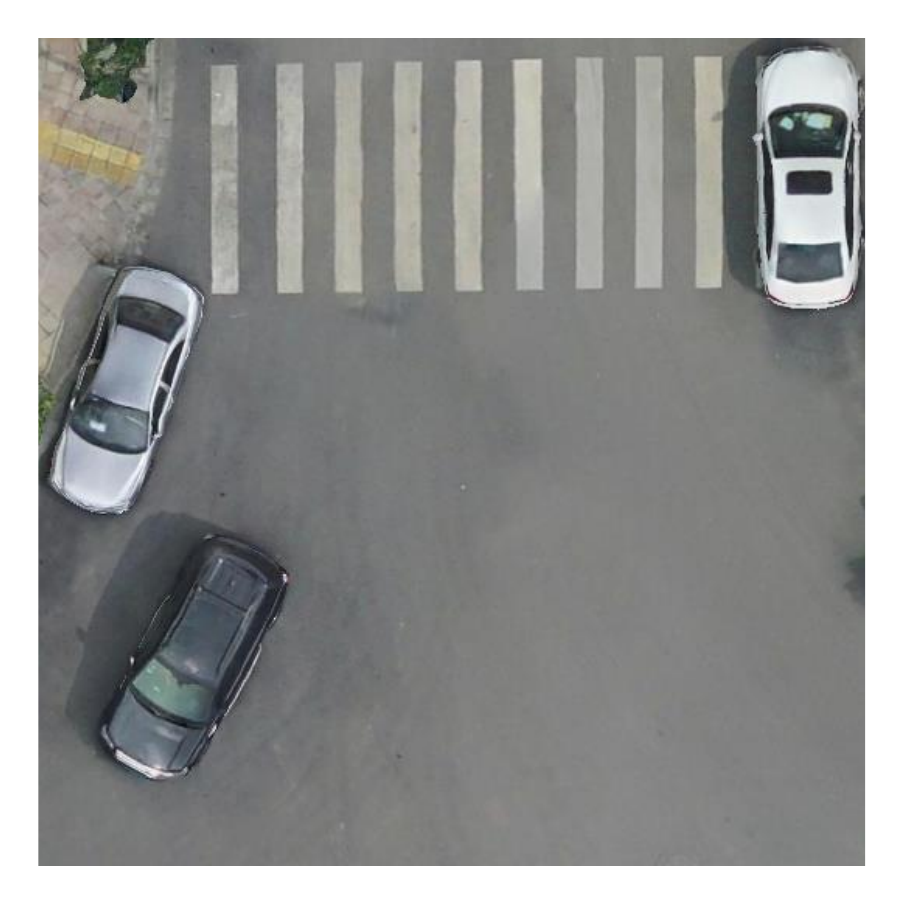}}
	\subcaptionbox{Detected objects}[0.32\linewidth]{\includegraphics[width=\linewidth]{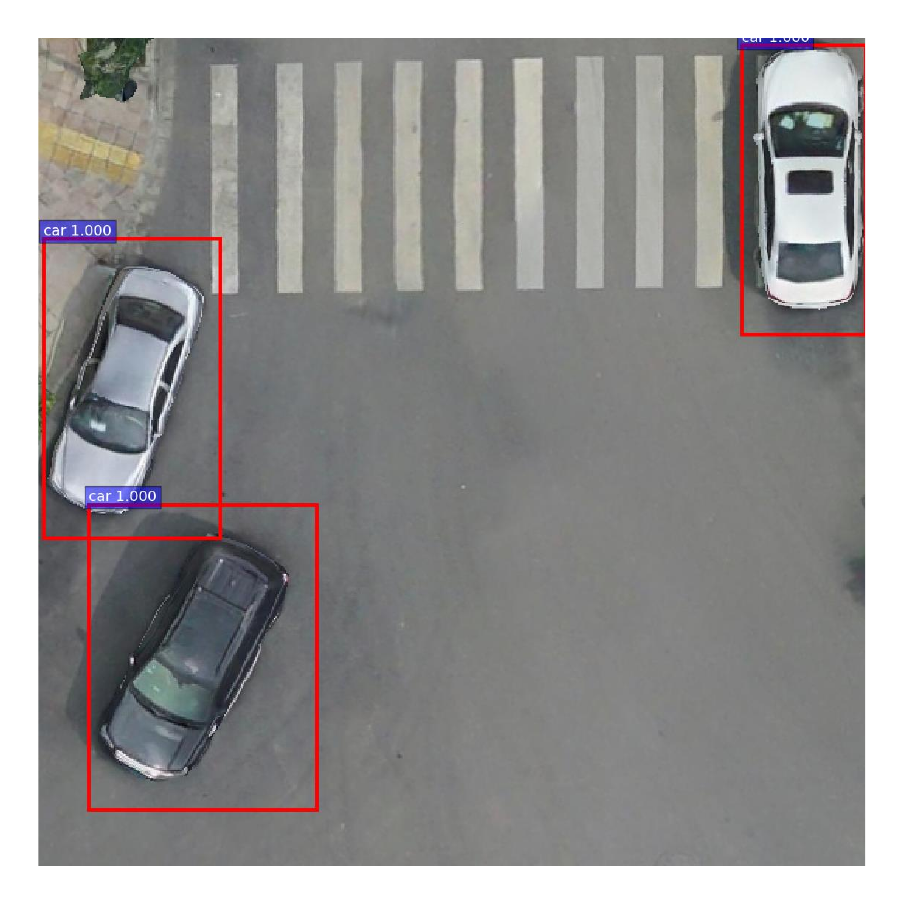}}
	\subcaptionbox{Dilated mask $\mathcal{R}_m$}[0.32\linewidth]{\includegraphics[width=\linewidth]{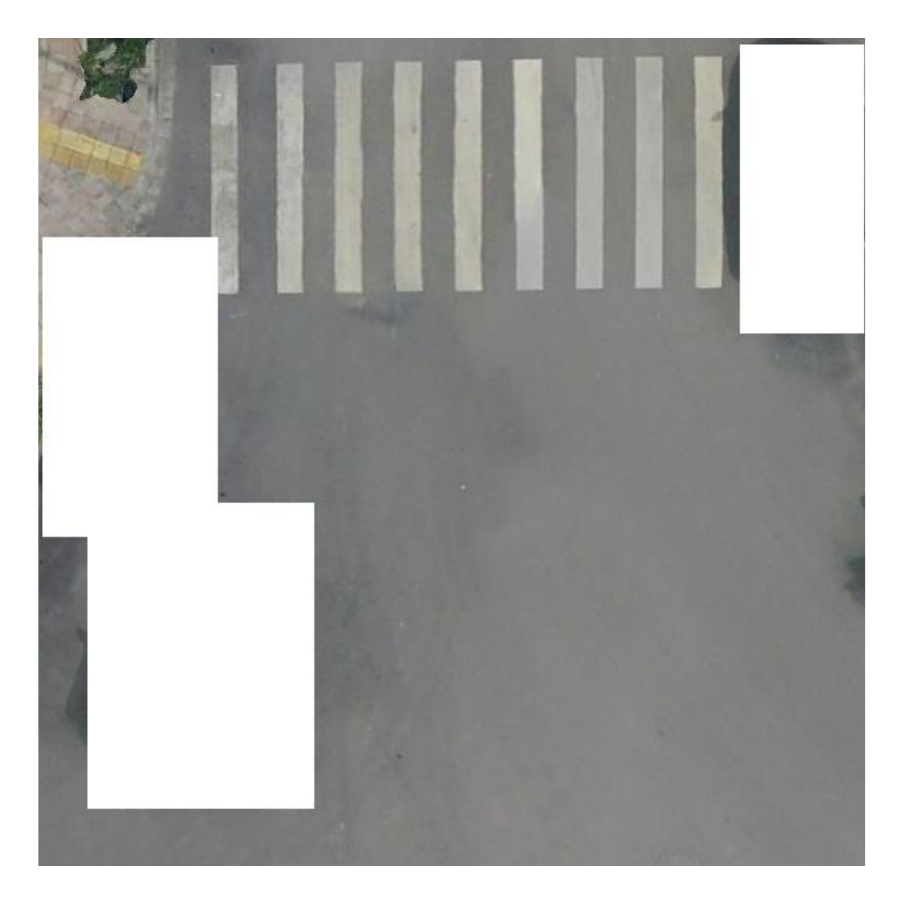}}
	\caption{Vehicle detection and mask generation. Faster R-CNN is used to detect objects in (a) the rendered color image, which are represented as (b) axis-aligned bounding boxes; the regions are dilated to account for shadows and other defects, and (c) are masked for further completion.}
	\label{fig:faster_rcnn}
\end{figure}

To remove the vehicles from the photogrammetric mesh models, we use  Faster R-CNN \citep{ren2015faster} for vehicle detection in the rendered image $\mathcal{R}_c$, as shown in Figure \ref{fig:faster_rcnn}.
As Faster R-CNN \citep{ren2015faster} is a well-established and industrially proven method, we only briefly introduce the training process in the following.

We use a pre-trained VGG16 model \citep{simonyan2014very} as the backbone for feature extraction owing to its simplicity.
Although existing datasets \citep{everingham2010pascal,lin2014microsoft} already contain training samples for vehicles, they are generally in perspective view rather than orthogonal view from the top.
Therefore, we interactively label a small set of datasets for object detection using LabelImg \citep{tzutalin2015labelimg}.
The datasets are exported in Pascal VOC format \citep{everingham2010pascal}.

We use photogrammetric mesh models covering both the campus of Southwest Jiaotong University (SWJTU) and part of Shenzhen to generate the training samples.
The images were obtained using an UAV (unmanned aerial vehicle) and a manned aircraft for SWJTU and Shenzhen, respectively.
Typical road areas are selected and rendered from the tiled mesh models.
In summary, approximately 250 patches from these two datasets are collected and interactively labeled for training.
During training, data augmentation by rotation and mirroring are used for improved generalizability.
After the detection of the bounding boxes, each object is enlarged by approximately 10\% to account for shadows and other possible defects in the textured models. The output is a masked raster $\mathcal{R}_m$, as shown in Figure \ref{fig:faster_rcnn}c.

\subsection{Extraction of linear regularities}
\label{subs:regularity_extraction}

Structured scenes, such as regularly arranged objects, are probably the most challenging cases for image completion \citep{he2012statistics}.
Unfortunately, scenes of urban roads featuring repeated line markers are highly structured.
It is desirable to explicitly consider regularities in the image completion process for improved performance in structured environments \citep{liu2010computational}.
Therefore, in this study, we first extract two types of linear regularities before the completion of the void regions in the rendered images, that is, translational regularities and linear features.

More specifically, this study adopts the PatchMatch-based \citep{barnes2009patchmatch} approach to complete void regions of vehicles based on \cite{huang2014image}.
PatchMatch is based on an effective approach to generate the nearest neighbor field (NNF) $\mathcal{N}$.
Each pixel of the NNF $\boldsymbol{v}=\mathcal{N}(\boldsymbol{p})$ denotes the offset $\boldsymbol{v}$ to the correspondence in the same image.
The completion is performed by filling the void region with the most self-similar patch.
The regularities are injected in this step to guide the generation of the NNF, which not only accelerates the convergence speed but also makes the NNF structure-aware.

\paragraph{Translational Regularity}
Inspired by previous work \citep{he2012statistics,huang2014image}, we also detect translational regularities using matched image features.
However, we found that for urban roads, the offsets between matched features exhibit clear linear regularities.
Specifically, we first extract SIFT \citep{lowe2004distinctive} features in the rendered image and then obtain matches using the standard ratio check.
Unlike in the case of feature matching between two images, no further outlier filtering using random sample consensus (RANSAC) \citep{fischler1981random} is used.

\begin{figure}[H]
	\centering
	\subcaptionbox{}[0.32\linewidth]{
		\includegraphics[width=\linewidth]{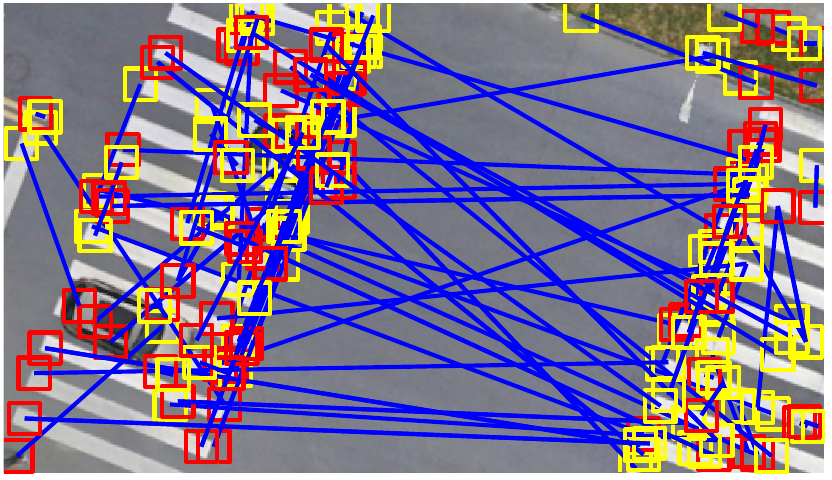}\vspace{1em}
		\includegraphics[width=\linewidth]{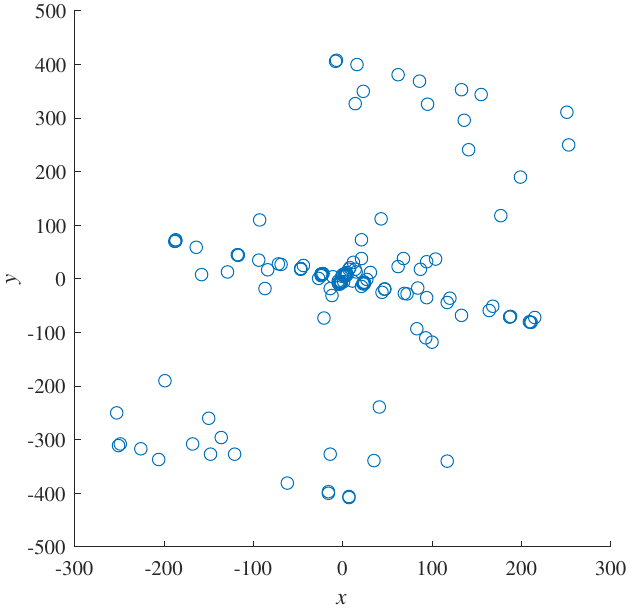}
	}
	\subcaptionbox{}[0.32\linewidth]{
		\includegraphics[width=\linewidth]{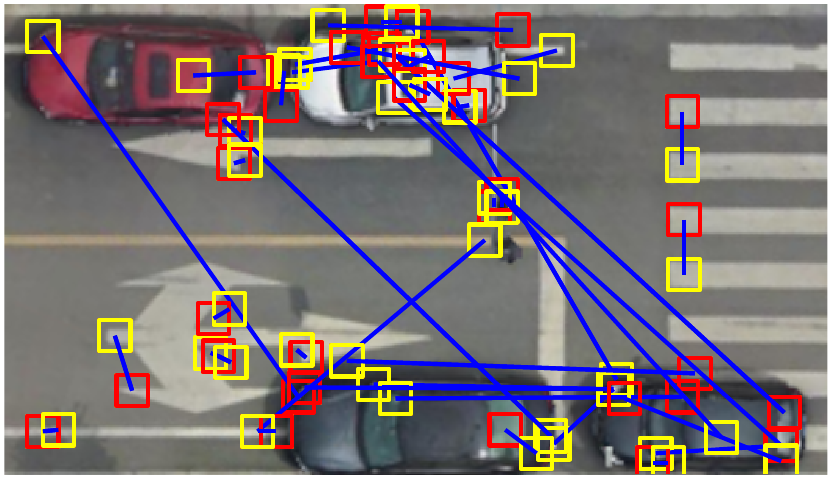}\vspace{1em}
		\includegraphics[width=\linewidth]{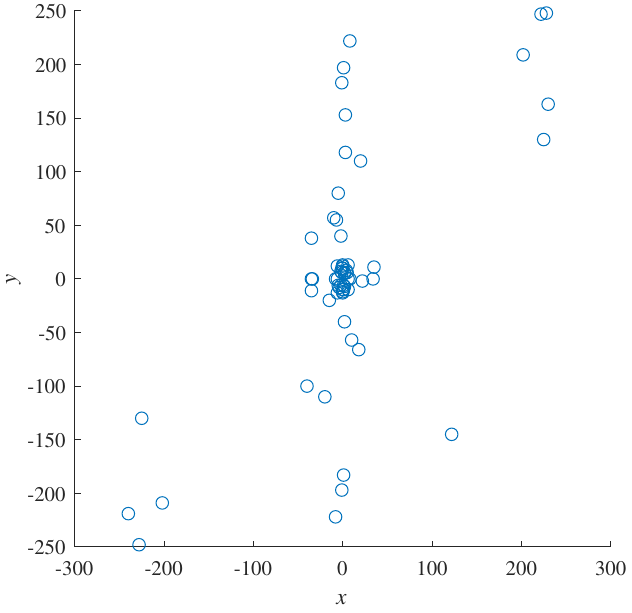}
	}
	\subcaptionbox{}[0.32\linewidth]{
		\includegraphics[width=\linewidth]{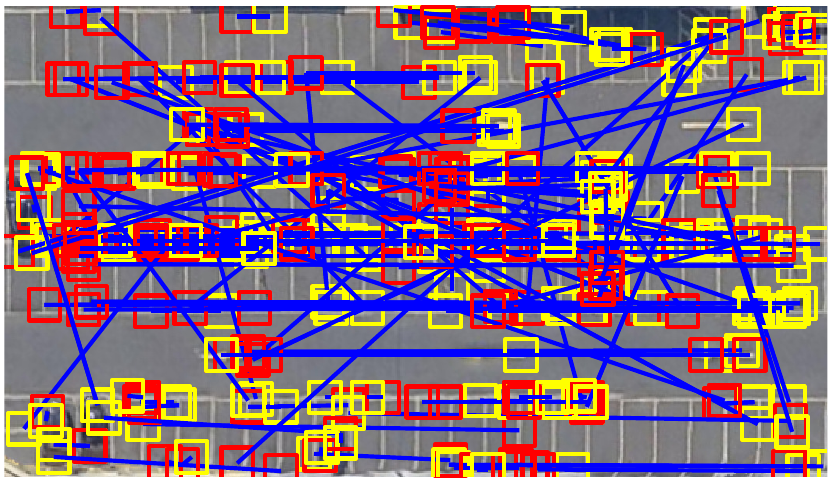}\vspace{1em}
		\includegraphics[width=\linewidth]{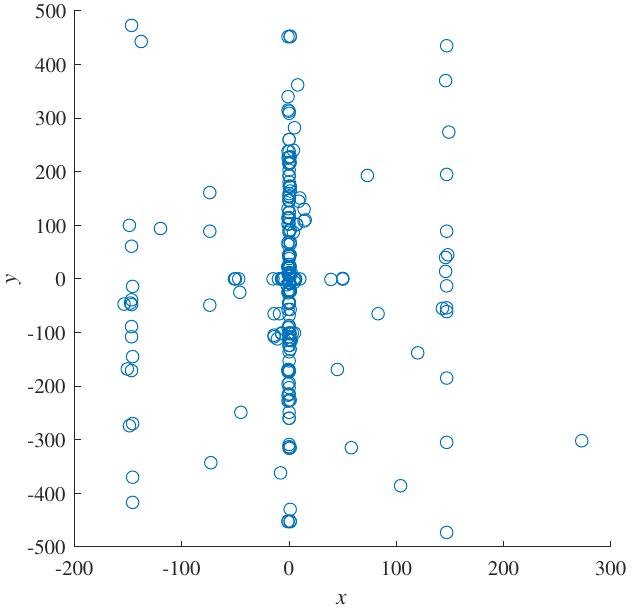}
	}
	\caption{Translational regularities of the matches for three typical scenarios. The top row shows the feature matches of the images to be completed, where the blue lines indicate the offsets of two matched points (red and yellow rectangles). The bottom row shows scatter plots of the offsets for the corresponding match results, which exhibit obvious linear patterns.}
	\label{fig:offsets}
\end{figure}

Three typical results for the feature matches are shown in Figure \ref{fig:offsets}.
It should be noted that the offsets of the feature matches exhibit a clear pattern of pointwise and linear clusters.
The pointwise clustering center indicates that similar patches are generally located at fixed intervals, such as equal distances between the road markings.
On the other hand, the linear pattern indicates that self-similar patches are quite likely to be retrieved by searching along the corresponding direction.
In addition, the orthogonal direction is commonly coexistent.
Although pointwise clusters are also highly common, in this study, we only enforce linear regularities for two reasons: (1) The point centers are generally aligned along the same line, and (2) as there are an excessive number of point centers, considering all of them significantly affects the efficiency of  patch-matching completion \citep{barnes2009patchmatch}. 
In summary, the orientations $\theta$ detected by RANSAC \citep{fischler1981random} and the orthogonal directions are then used in structure-aware image completion, that is, a set of $n$ angles $\Theta=\{\theta_1,...,\theta_n\}$.

\begin{figure}[H]
	\centering
	\subcaptionbox{LSD}[0.3\linewidth]{
		\includegraphics[width=\linewidth]{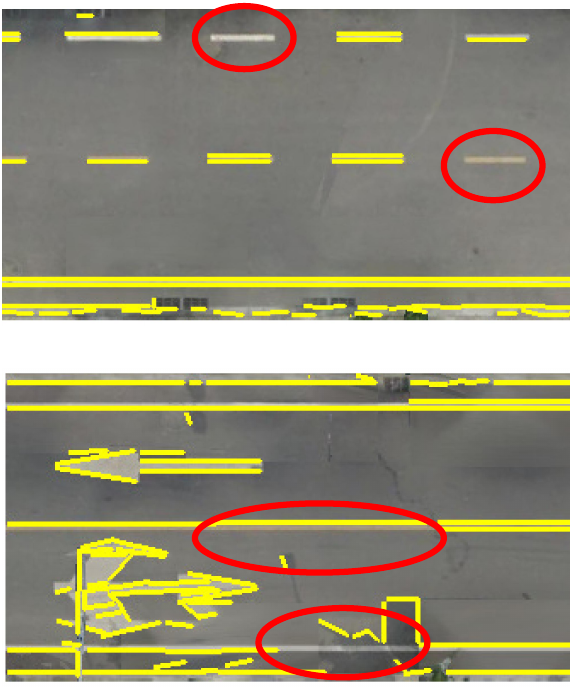}
	}
	\subcaptionbox{Prewitt}[0.3\linewidth]{
		\includegraphics[width=\linewidth]{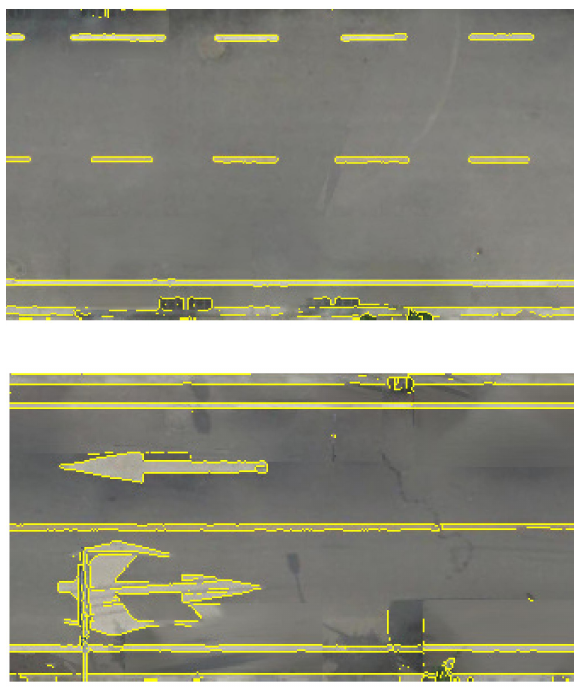}
	}
	\subcaptionbox{Canny}[0.3\linewidth]{
		\includegraphics[width=\linewidth]{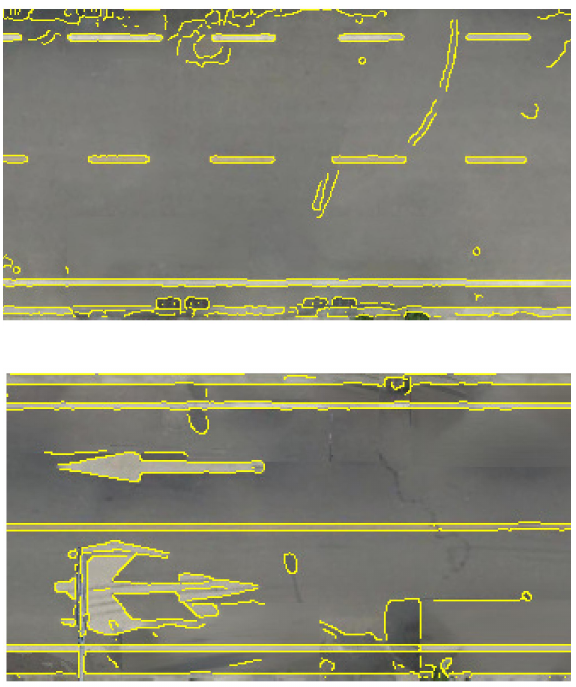}
	}
	\caption{Detected edges by different methods. (a) Lines extracted by LSD. (b) and (c) show the detected edges using Prewitt and Canny operators, respectively.}
	\label{fig:edge_detection}
\end{figure}

\paragraph{Edge Feature}
Translational regularities are a good indicator of mid-level knowledge for the scenes, and they are used in guided completion.
In addition, we directly consider low-level vision features, such as image edges, which also form the skeleton of road scenes.
Although there are several choices to detect edge features, such as the line segment detector (LSD) \citep{von2012lsd,zhu2020interactive}, we found that LSD and other approaches designed for contour extraction are more sensitive to scene noise (shown in Figure \ref{fig:edge_detection}).
Unfortunately, the rendered image from the textured meshes is inevitably noise-laden owing to defects in the MVS pipeline.
Therefore, we directly adopt an efficient gradient filter to extract binary edges $\mathcal{R}_l$ from the rendered image $\mathcal{R}_c$, namely, the Prewitt filter.

\subsection{Structure-aware completion of urban roads}
\label{s:image_completion}

\begin{figure}[h]
	\centering
	\includegraphics[width=\linewidth]{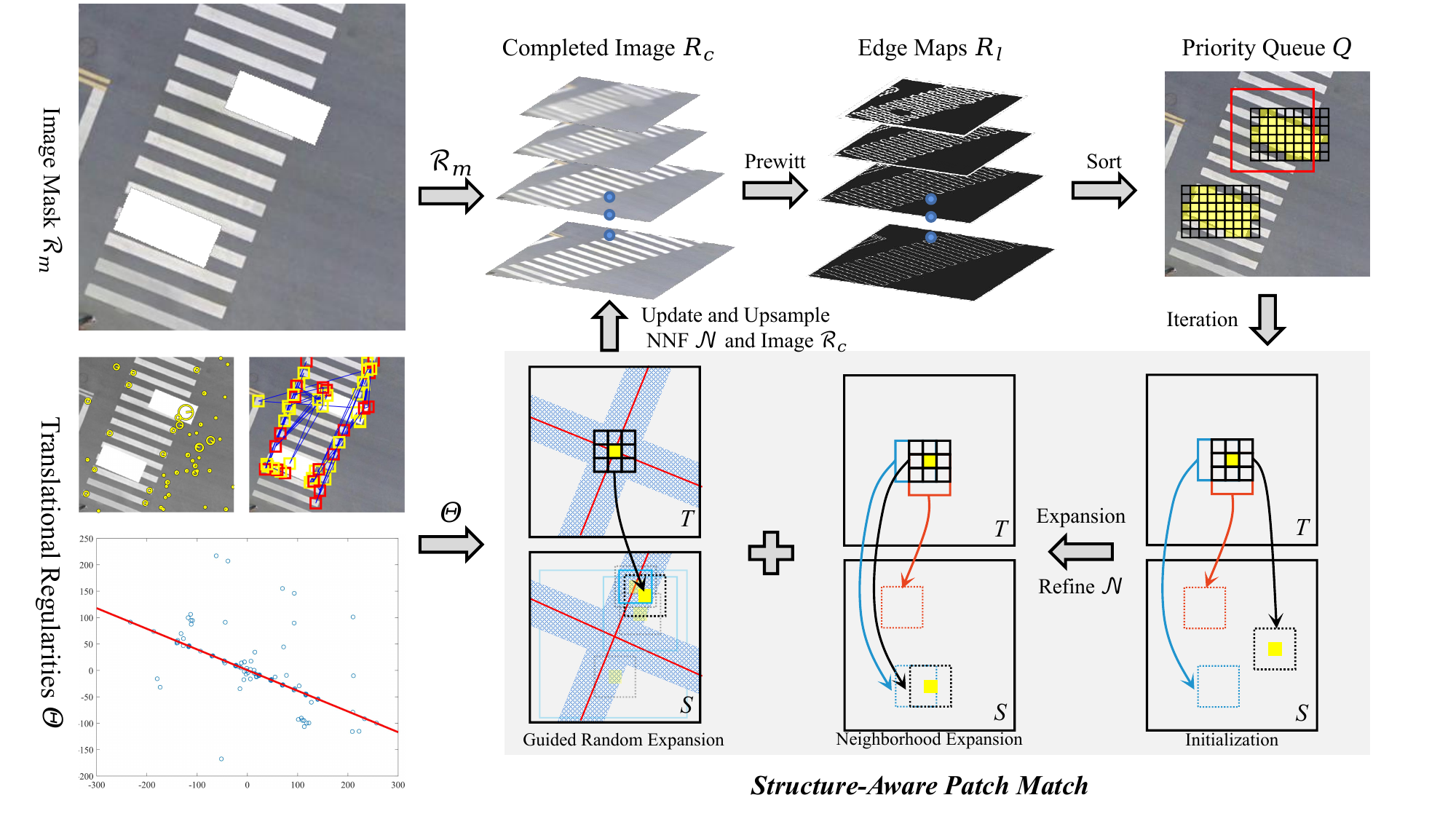}
	\caption{Pyramid scheme for the structure-aware completion of urban roads. A priority queue is generated from the edge maps to prioritize the structured region. In addition, a guided random expansion is considered with translational regularities to retrieve more structured regions. The NNF $\mathcal{N}$ determines the offset between the most self-similar source $S$ and the target $T$ patches.}
	\label{fig:image_completion}
\end{figure}

\begin{figure}[H]
	\centering
	\includegraphics[width=\linewidth]{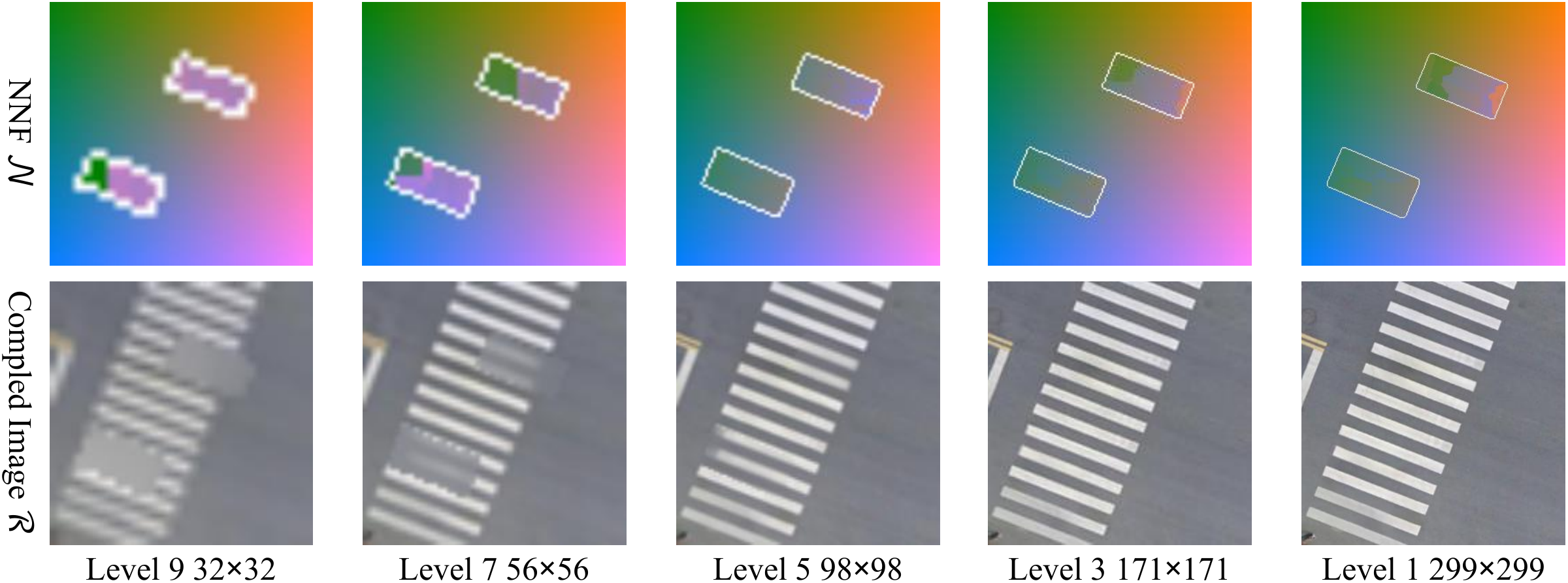}
	\caption{Updated NNF $\mathcal{N}$ and completed image $\mathcal{R}_c$ using a pyramid scheme. The white masks in the first row indicate the masked regions.}
	\label{fig:NNF}
\end{figure}

The objective of the completion of urban roads is to recover void regions in $\mathcal{R}_m$ (Figure \ref{fig:faster_rcnn}c) using the orientations $\Theta$ of translational regularities (Figure \ref{fig:offsets}) and edge maps $\mathcal{R}_l$.
A pyramid scheme is established based on the PatchMatch strategy \citep{barnes2009patchmatch} to search for the best NNF $\mathcal{N}$ progressively (Figure \ref{fig:image_completion}).
Namely, beginning with the coarsest level, a random $\mathcal{N}$ is initialized, and the edge map $\mathcal{R}_l$ is generated from the coarsest image mask $\mathcal{R}_m$.
We first establish a priority queue $Q$, which prefers pixels with higher edge scores. The order of the PatchMatch-based expansion is determined by the priority queue $Q$ rather than the original scanline-based strategy \citep{barnes2009patchmatch}.
For each pixel $\boldsymbol{p}$ in $Q$, the NNF $\mathcal{N}(\boldsymbol{p})$ is refined using an improved updating strategy guided by the linear regularities $\Theta$.
An example of the progressively refined NNF $\mathcal{N}$ and completed image $\mathcal{R}_c$ is shown in Figure \ref{fig:NNF}.

\subsubsection{Generation of priority queue with edge maps}
\label{subs:priority}

Beginning with a randomly initialized NNF $\mathcal{N}$, the vanilla PatchMatch refines $\mathcal{N}$ in the canonical scanline direction in odd iterations, that is, top to down and left to right, and in the reverse direction in even iterations \citep{barnes2009patchmatch}.
However, this may clear artifacts in structured regions.
If the patch is expanded from the textureless region to the structured region, the regularity embedded in the structures may not be preserved.

Therefore, we propose an effective approach to remedy this.
We argue that structured regions generally have higher edge responses, that is, number of edge pixels in a local patch.
Specifically, the sum of the center-aligned square patch on the binary edge map $\mathcal{R}_l$ is used to represent the responses of the edge.
Then, the pixels in the void region are sorted in descending order according to the edge responses in the priority queue $Q = \{\boldsymbol{p}_1, \boldsymbol{p}_2, ..., \boldsymbol{p}_n\}$.

\subsubsection{Structure-aware similarity measure of patches}
To refine the NNF $\mathcal{N}$, \cite{barnes2009patchmatch} compares the similarity of the current pixel $\boldsymbol{p}$ with that of its 4-neighborhood with regard to the offset $\boldsymbol{v}=\mathcal{N}(\boldsymbol{p})$. Two square patches are generated as follows: 
\begin{equation}
	\begin{split}
		T(\boldsymbol{p})&=\{\mathcal{R}_c(\boldsymbol{p}+\boldsymbol{s})|\boldsymbol{s}\in[-\tfrac{W}{2},\tfrac{W}{2}]\times[-\tfrac{W}{2},\tfrac{W}{2}] \} \\
		S(\boldsymbol{p}, \boldsymbol{v})&=T(\boldsymbol{p}+\boldsymbol{v})
	\end{split}
\end{equation}
where $T$ and $S$ are the target and source patches centered around $\boldsymbol{p}$ and $\boldsymbol{p}+\boldsymbol{v}$, respectively.
The target patch $T$ is inside the region to be completed, and the source patch $S$ is outside the region.
It should be noted that during patch expansion, the offset $\boldsymbol{v}$ may not be sampled from the corresponding pixel $\boldsymbol{p}$ of NNF $\mathcal{N}$; it can also be sampled from the 4-neighborhood $N_4(\boldsymbol{p})$ of $\boldsymbol{p}$ or even from a random location \citep{barnes2009patchmatch}.
\red{$W$ is the patch size. Considering the effect and efficiency, this paper sets $W$ to 21.} 
The similarity (or matching cost) can be intuitively computed from the grayscale values of the two patches \citep{barnes2009patchmatch}.
However, we also consider the regularities $\Theta$ in the computation of the similarity measure.
Specifically, the similarity measure consists of three terms: appearance $E_a$, proximity $E_p$, and regularity $E_r$. That is,
\begin{equation}
	E=E_a+\lambda_1 E_p + \lambda_2 E_r
\end{equation}
\red{where $\lambda_1=5\times10^{-4}$ and $\lambda_2=0.5$ are chosen \citep{huang2014image}.}

\paragraph{1) Appearance cost $E_a(\boldsymbol{p},\boldsymbol{v})$.}
We use the sum of the absolute differences between the target and source patches to measure the appearance difference as follows: 
\begin{equation}
	E_a(\boldsymbol{p},\boldsymbol{v})=\sum_{i} w_i| T_i(\boldsymbol{p}) - T_i(\boldsymbol{p}+\boldsymbol{v}))|
\end{equation}
where $T_i$ and $S_i$ denote the grayscale values of the $i$-th pixel of the patch,
and $w_i$ is an isotropic weight generated from a Gaussian kernel \citep{huang2014image}.

\paragraph{2) Proximity cost $E_p(\boldsymbol{p,v})$}
Generally, nearby pixels are preferred over distant pixels \citep{kopf2012quality}. Therefore, we add an additional penalty $E_p$ to prevent selecting distant patches in the NNF $\mathcal{N}$, as in \citep{huang2014image}:
\begin{equation}
	E_p(\boldsymbol{p,v})=\frac{||\boldsymbol{v}||^2}{\sigma_d(\boldsymbol{p})^2+\sigma_c^2}
\end{equation} 
where $\sigma_d$ and $\sigma_c$ are normalizers. Specifically, $\sigma_d(\boldsymbol{p})$ is the distance to the nearest border of the invalid regions, and $\sigma_c=\max(w,h)/8$ accounts for the size of image $(w,h)$.
$E_p(\boldsymbol{p})$ prefers small offsets in the NNF $\mathcal{N}$.

\paragraph{3) Regularity cost $E_r(\boldsymbol{v})$.}
In structured scenes, the offset direction $\theta_{\boldsymbol{v}}$ in the NNF should be consistent with the detected regularities $\Theta$. We use the minimum angle difference to measure the regularity cost.
\begin{equation}
	E_r(\boldsymbol{v}) = \min_{\theta \in \Theta}{\cos(\theta_{\boldsymbol{v}} - \theta)}
\end{equation}

\subsubsection{Guided random expansion}
\label{subs:direction_constraint}

During the expansion of the NNF $\mathcal{N}$, PatchMatch \citep{barnes2009patchmatch} tests the similarities of both the 4-Neighborhood $N_4(\boldsymbol{p})$ and a set of random pixels centered around $\boldsymbol{p}$ as $R(\boldsymbol{p})=\{\boldsymbol{q}_1,\boldsymbol{q}_2, ..., \boldsymbol{q}_r\}$.
The random set prevents the occurrence of local minima of the expansion.
In this study, the directions determined by the translational regularities $\Theta$ are also used to guide the random expansion, as in the case of the regularity cost.
Specifically, the random pixels should be selected in a rectangular buffer along the direction $\theta \in \Theta$ and the orthogonal direction $\theta + \tfrac{\pi}{2}$, as indicated by the shaded blue region in Figure \ref{fig:image_completion}. 
In addition, a series of $r$ random pixels with decreasing radius are selected.
The radius for pixel $\boldsymbol{q}_r$ is determined by $\max(w,h) \times (\tfrac{1}{2})^r$.
In summary, a single iteration of the refinement process for the proposed structure-aware PatchMatch is shown in Algorithm \ref{algo:patch-match}.

\begin{algorithm}[H]
	\caption{Structure-aware PatchMatch.}
	\label{algo:patch-match}
	\begin{algorithmic}[2]
		\Procedure{PatchMatch}{$\mathcal{N},\mathcal{R}_c,Q,R$}
		\For{$\boldsymbol{p} \in Q$ }

		\For{$\boldsymbol{q} \in N_4(\boldsymbol{p})$} \Comment{Neighborhood Expansion}
			\If{$E(\boldsymbol{p}, \mathcal{N}(\boldsymbol{q})) < E(\boldsymbol{p}, \mathcal{N}(\boldsymbol{p}))$}
				\State $\mathcal{N}(\boldsymbol{p}) \gets \mathcal{N}(\boldsymbol{q})$
			\EndIf
		\EndFor
		
		\For{$\boldsymbol{q} \in R(\boldsymbol{p})$} \Comment{Guided Random Expansion}
			 \If{$E(\boldsymbol{p}, \mathcal{N}(\boldsymbol{q})) < E(\boldsymbol{p}, \mathcal{N}(\boldsymbol{p}))$}
			 \State $\mathcal{N}(\boldsymbol{p}) \gets \mathcal{N}(\boldsymbol{q})$
			 \EndIf
		\EndFor
		
		\EndFor
		\State \Return $\mathcal{N}$
		\EndProcedure
	\end{algorithmic}
\end{algorithm}

\section{Experimental evaluation and analysis}
\label{s:results}

\subsection{Dataset description}
\label{subs:data}

\begin{figure}[h]
	\centering
	\subcaptionbox{SWJTU}[\linewidth]{
		\includegraphics[width=0.48\linewidth]{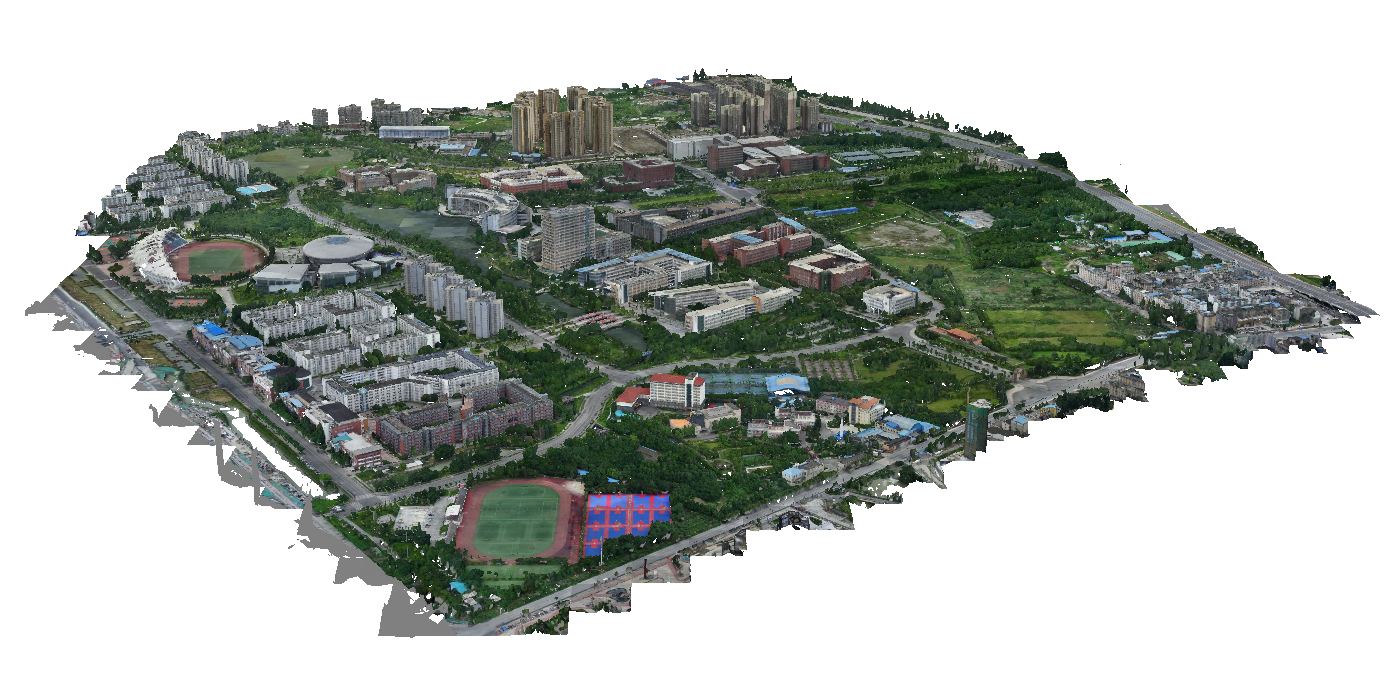}
		\includegraphics[width=0.25\linewidth]{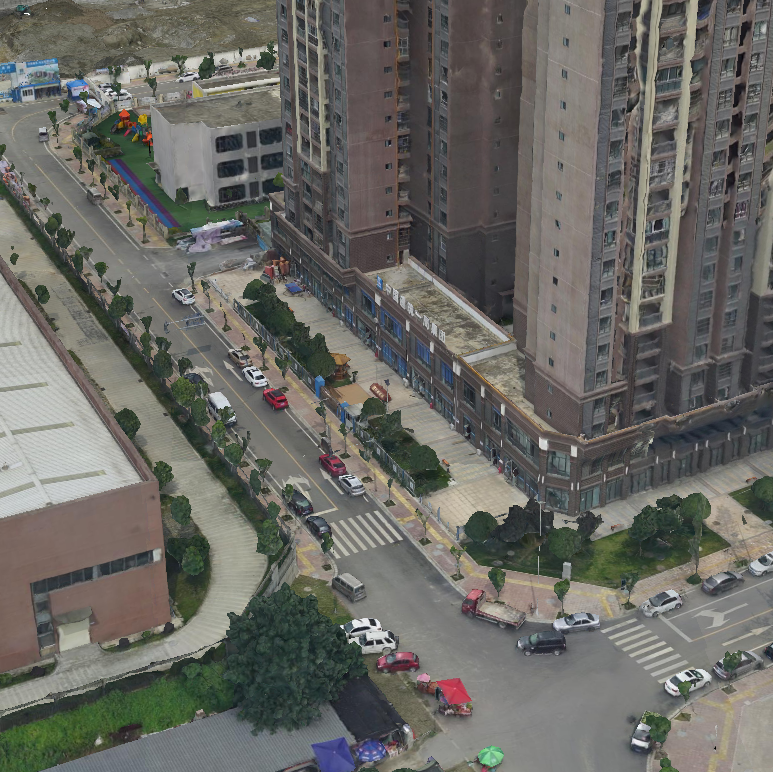}
		\includegraphics[width=0.25\linewidth]{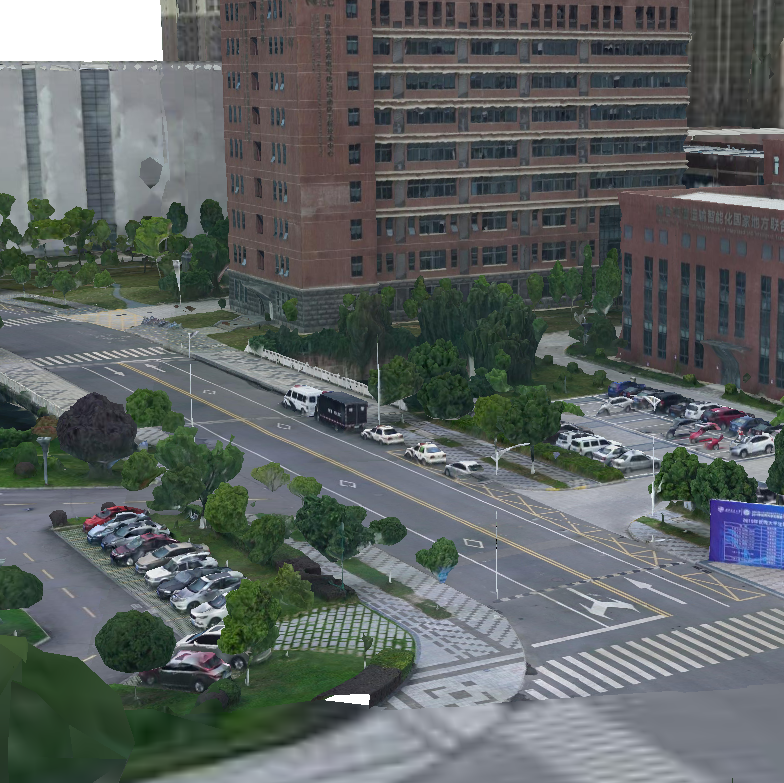}
	}
	\subcaptionbox{Shenzhen}[\linewidth]{
		\includegraphics[width=0.48\linewidth]{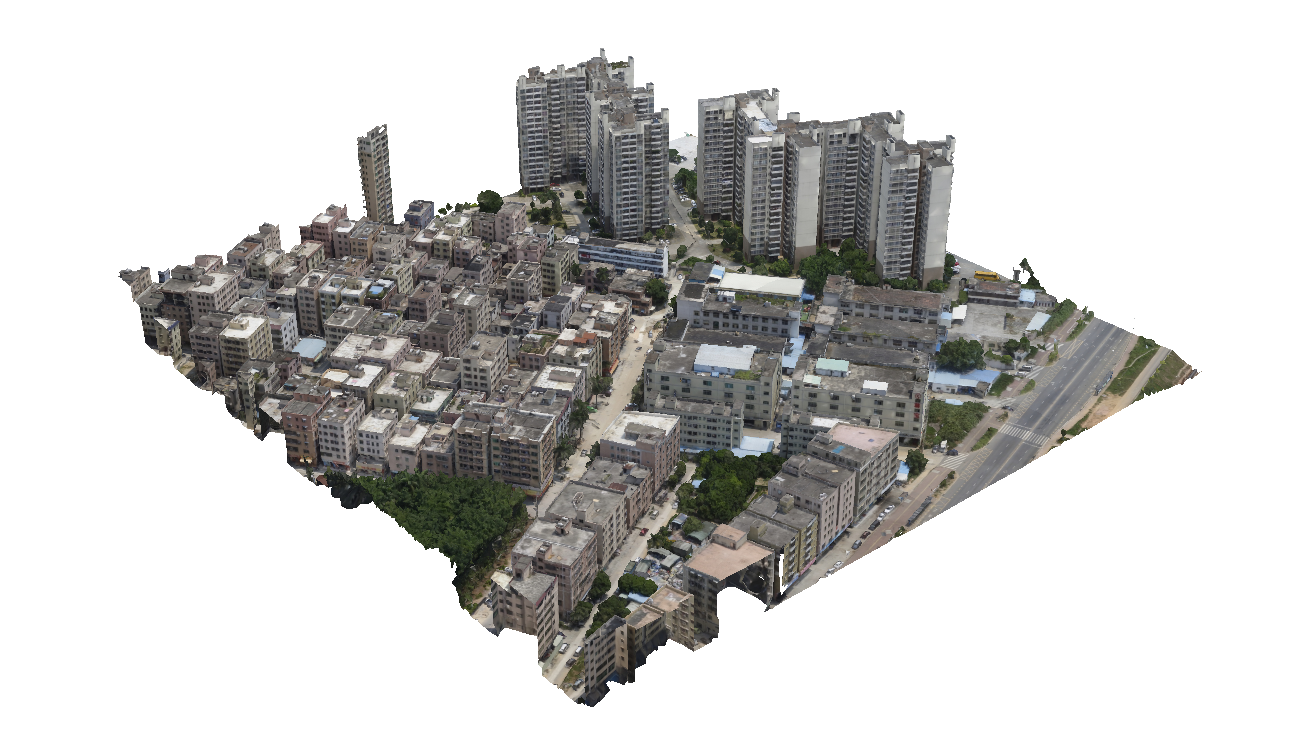}
		\includegraphics[width=0.25\linewidth]{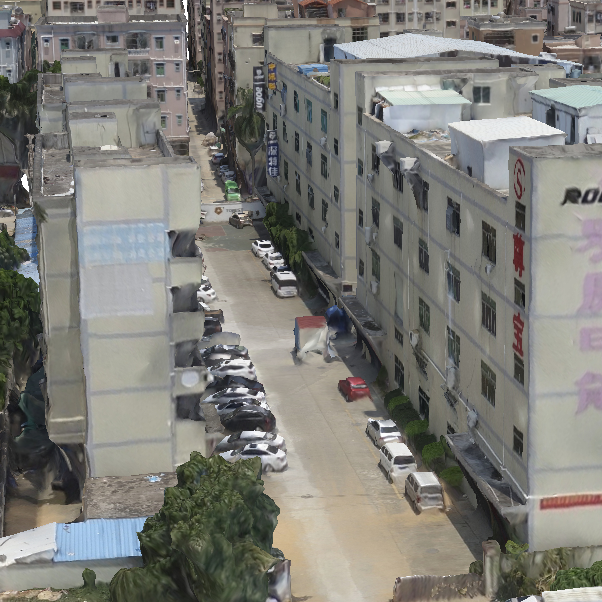}
		\includegraphics[width=0.25\linewidth]{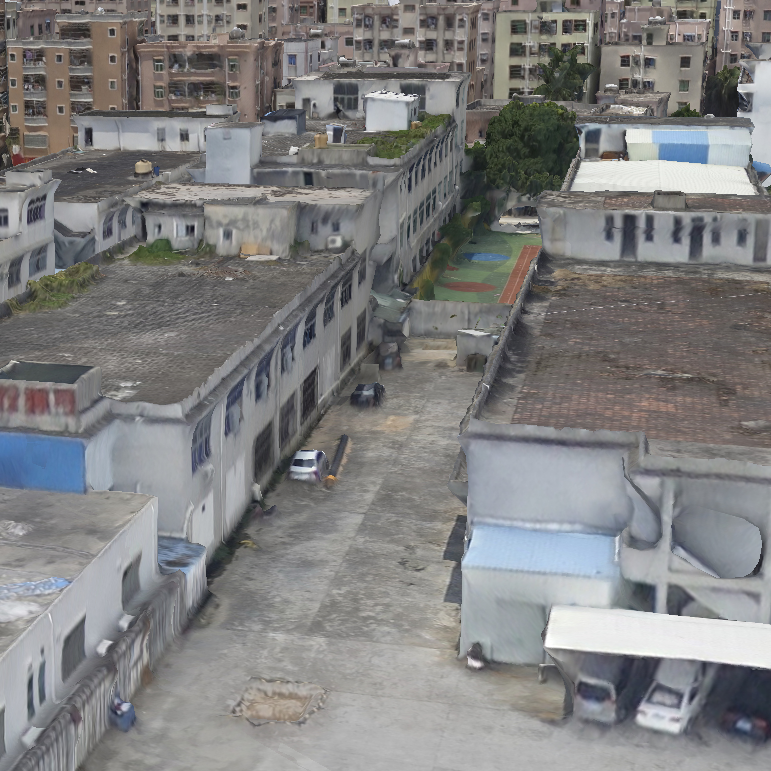}
	}
	\subcaptionbox{Dortmund}[\linewidth]{
		\includegraphics[width=0.48\linewidth]{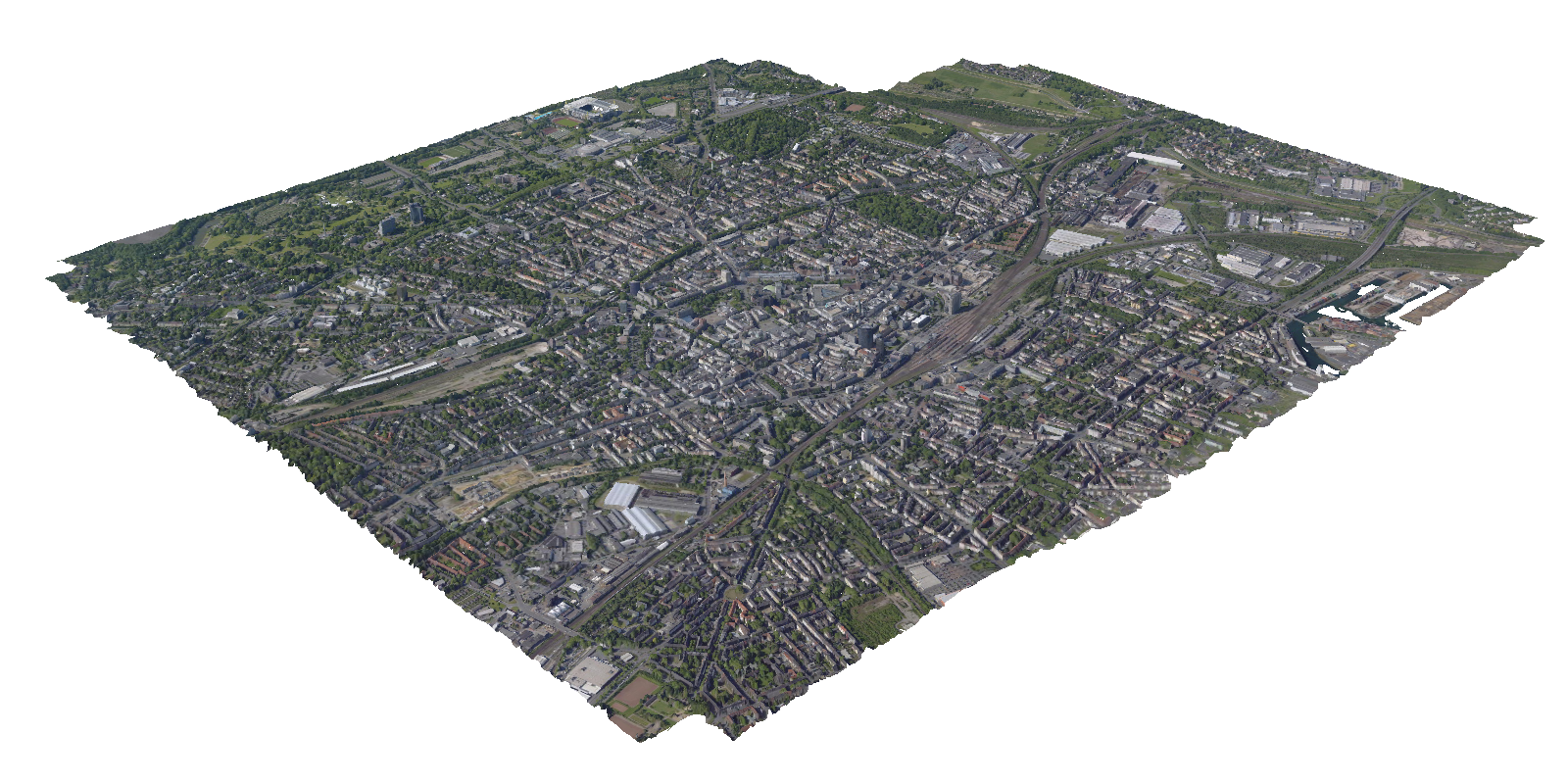}
		\includegraphics[width=0.25\linewidth]{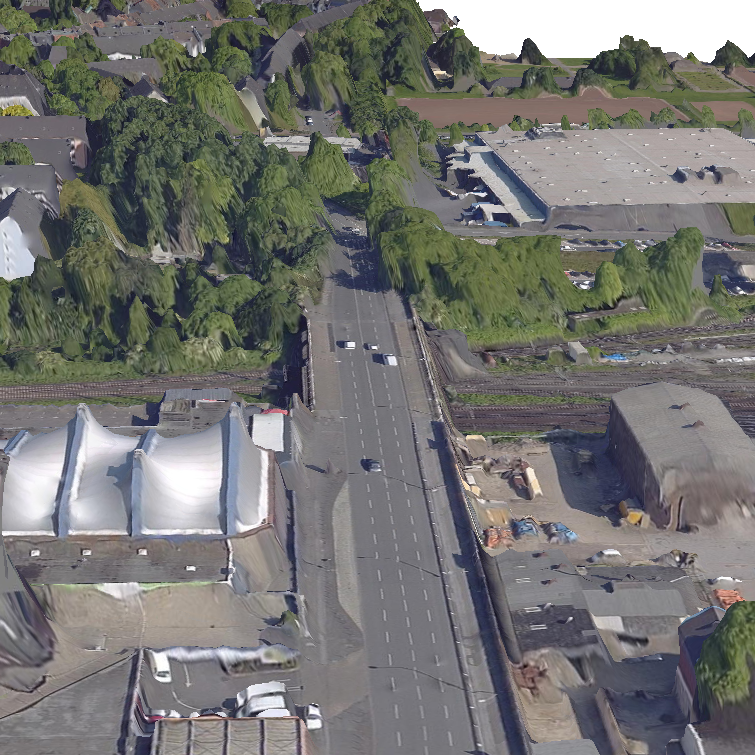}
		\includegraphics[width=0.25\linewidth]{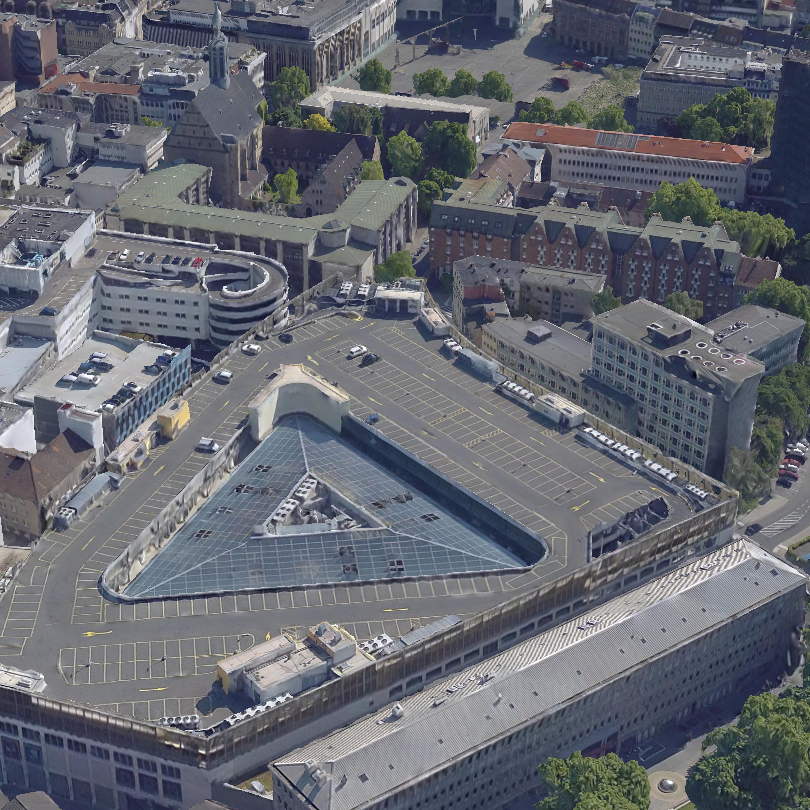}
	}
	\caption{Datasets used for the experimental evaluations. Different road scenarios are considered. In addition, the three datasets were captured by different sensors at different ground resolutions.}
	\label{fig:dataset}
\end{figure}

To evaluate the proposed methods, we use three datasets, which consist of highways, parking lots, and roads with and without marker lines (Figure \ref{fig:dataset}). 
The first was collected from the campus of Southwest Jiaotong University (SWJTU) in Chengdu, China.
Two typical regions are considered, including a road with no parking signs and a cross intersection.
The second was collected from a block in Shenzhen of China.
We select an alley with numerous vehicles parked and a residential area for the experiments.
The third was provided by courtesy of ISPRS in Dortmund, Germany \citep{nex2015isprs}.
The road areas on a bridge and a park lot on a building roof were selected for the experiments.

In addition to the different road scenarios, the three datasets were also captured by different sensors at different relative flight heights on different platforms.
Table \ref{tab:datasets} lists the type of sensor, relative flight height, ground sample distance (GSD), and the number of images for the three datasets.
It should be also noted that the SWJTU dataset was captured by a UAV, and the other two by a manned aircraft.

\begin{table}[h]
\centering
\caption{Detailed description of the datasets}
\label{tab:datasets}
	\begin{tabular}{c|l|ccc}
		\hline
		\multicolumn{2}{c|}{Dataset}                     & SWJTU          & Shenzhen  & Dortmund     \\ \hline
		\multicolumn{2}{c|}{Sensor}                      & SONY ICLE-5100 & PHASE ONE IQ180 & IGI PentaCam \\
		\multicolumn{2}{c|}{Relative flight height (m)}  & 85.23              & 918.66    & 831.97            \\
		\multicolumn{2}{c|}{Ground sample distance (cm)} & 1.6-2.0        & 6.0-8.0         & 8.0-12.0     \\ \hline
	\end{tabular}
\end{table}

\subsection{Results}
\label{subs:qualitative}

\subsubsection{Texture integration}
\begin{figure}[H]
	\centering
	\subcaptionbox{SWJTU}[0.32\linewidth]{\includegraphics[width=\linewidth]{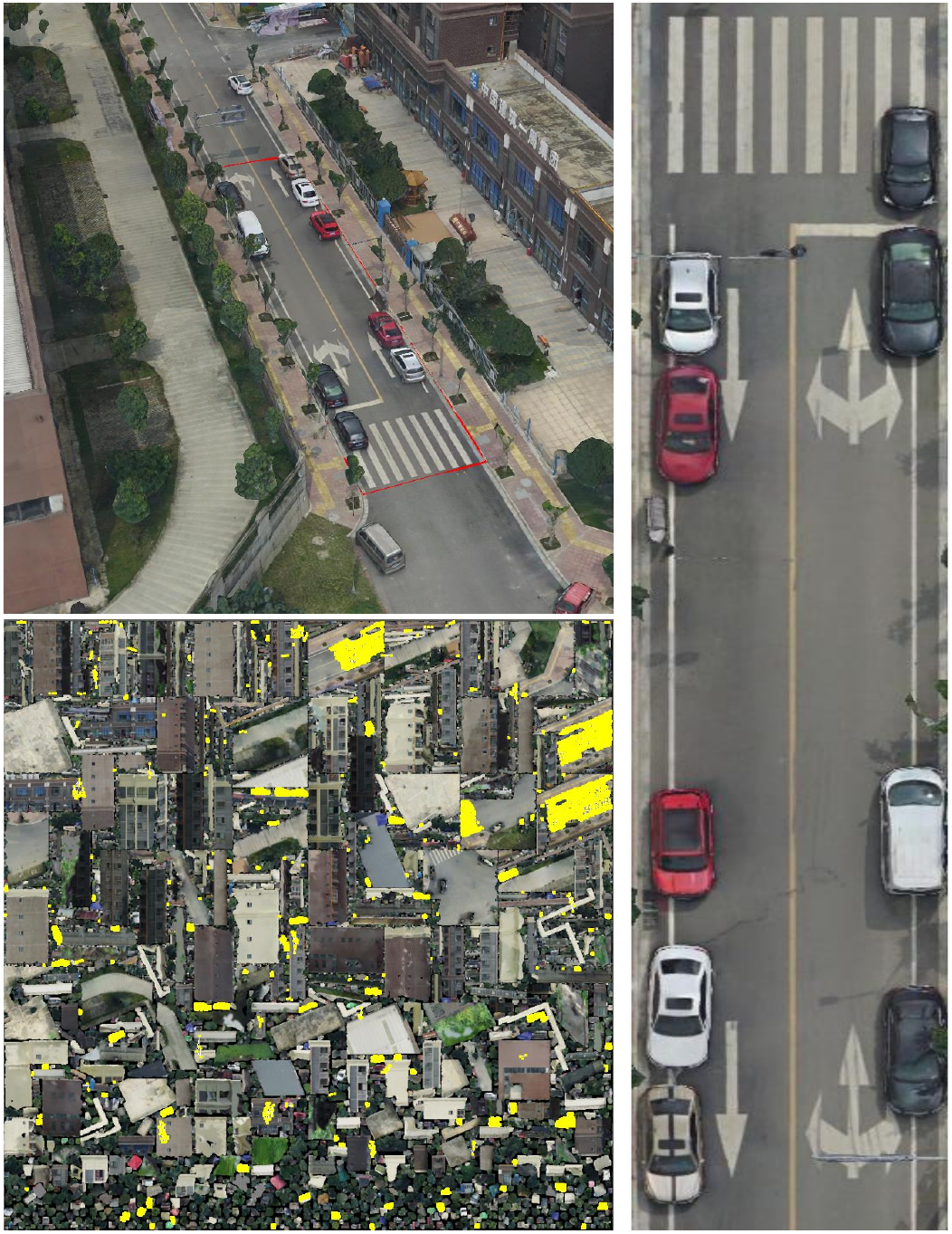}}
	\subcaptionbox{Shenzhen}[0.32\linewidth]{\includegraphics[width=\linewidth]{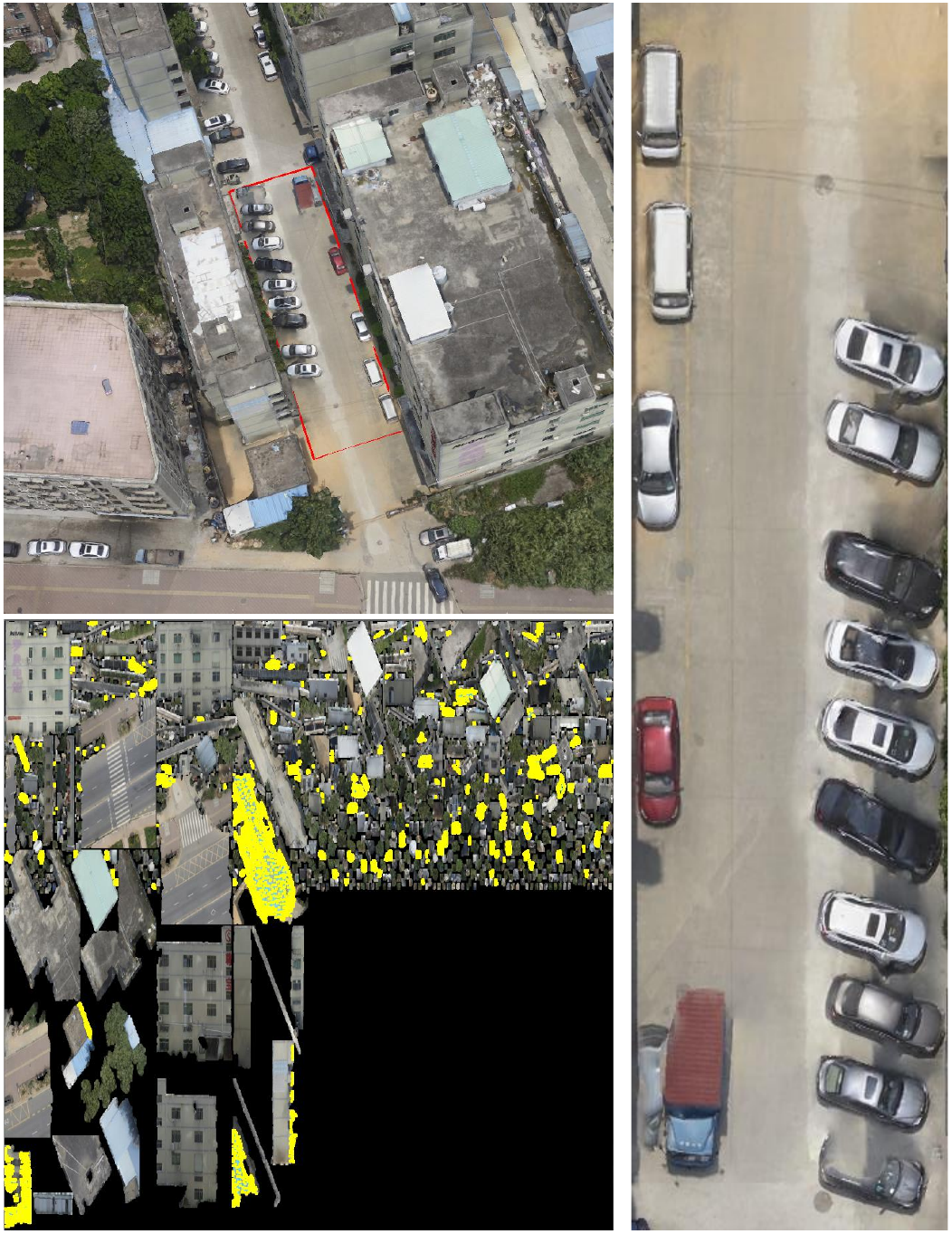}}
	\subcaptionbox{Dortmund}[0.32\linewidth]{\includegraphics[width=\linewidth]{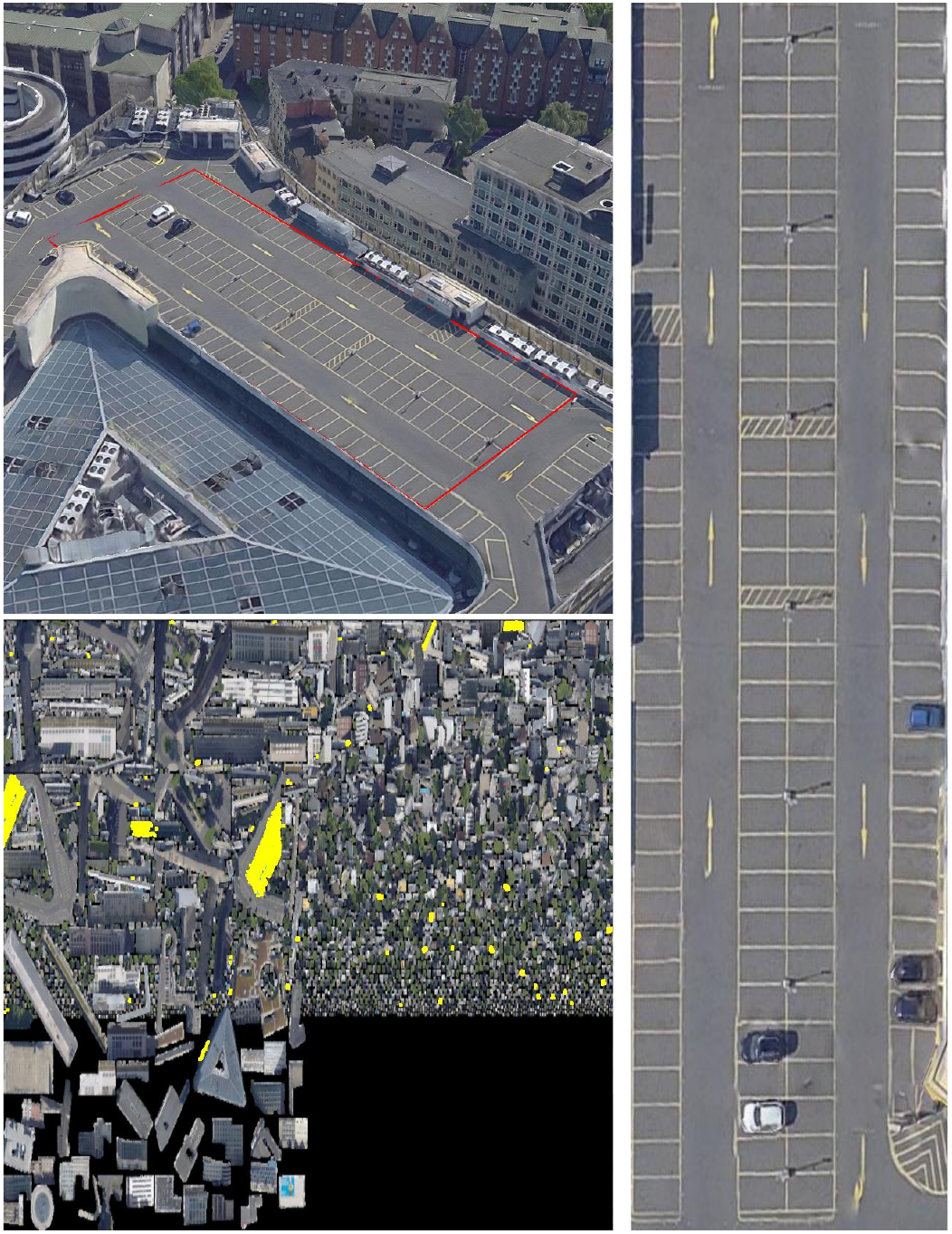}}
	\caption{Integration of the texture by orthogonal rendering of the mesh models. The red rectangle in the top-left of each subfigure is the selected ROI in the mesh model. The yellow highlight in the bottom-left part indicates the corresponding facets selected on the UV mesh. The right part of each subfigure shows the rendered image.}
	\label{fig:res_integration}
\end{figure}

To resolve the issue caused by the discontinuous texture image, we first integrate the texels in different charts into a unified image through orthogonal rendering of the textured mesh models.
Figure \ref{fig:res_integration} shows the integrated image for the three datasets.
The red rectangle in the top-left part of each subfigure is interactively selected by the operator, and only the triangles inside the selected ROI are considered for further processing.
The yellow highlighted regions in the bottom-left part of each subfigure indicate the facets involved in the UV mesh.
Even though the ROI only accounts for a small rectangular area, several small charts are involved; direct processing is difficult owing to the fragments in the UV mesh.
This is resolved through efficient rendering of the mesh models (Figure \ref{fig:res_integration}, right part).
The resolution of the viewport is selected according to the average GSD of the model; this ensures that the texture information is preserved during the rendering process.

\subsubsection{Image and mesh completion}
After obtaining the rendered image $\mathcal{R}_c$, we directly detect the vehicles using Faster R-CNN \citep{ren2015faster}, from which the mask image $\mathcal{R}_m$ is obtained.
Then, the void regions are completed using the proposed methods as $\mathcal{R}_c'$ and deintegrated to mesh texture.
In addition, we flatten the triangles inside the masked regions for the completed mesh $\mathcal{M}'$.

\begin{figure}[H]
	\centering
	\subcaptionbox{Image completion}[0.26\linewidth]{
		\includegraphics[width=\linewidth]{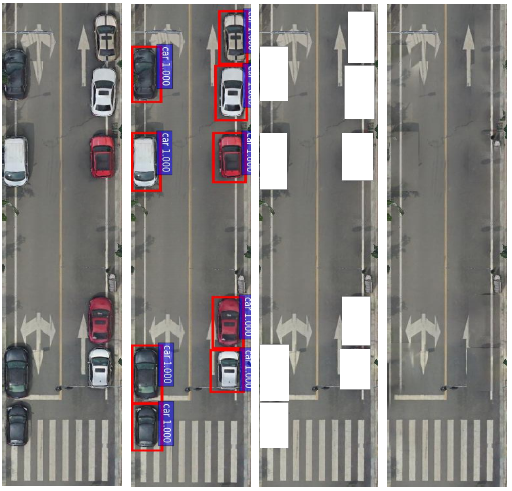}\vspace{0.2em}
		\includegraphics[width=\linewidth]{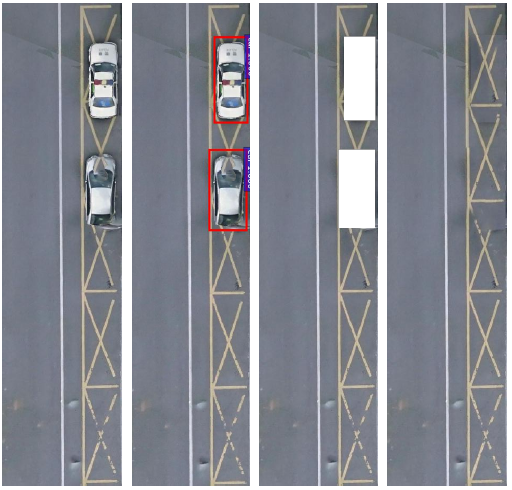}
	}
	\subcaptionbox{Mesh completion}[0.5\linewidth]{
		\includegraphics[width=\linewidth]{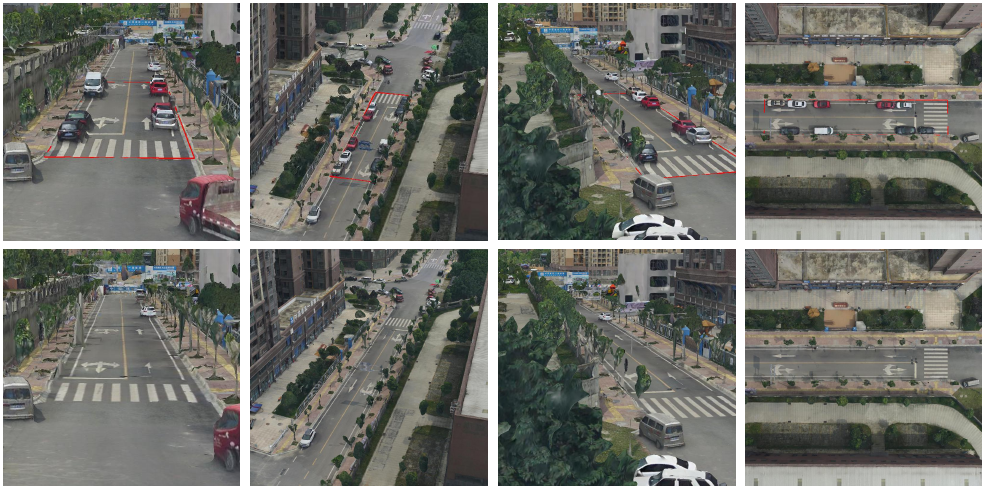}\vspace{0.2em}
		\includegraphics[width=\linewidth]{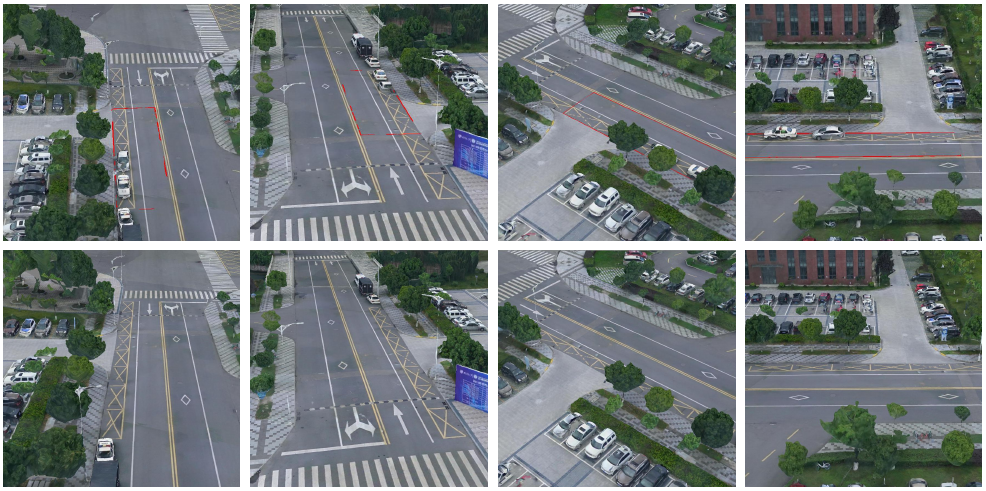}
	}
	\caption{Completed mesh models for the two regions of the SWJTU dataset. (a) Completion of images; the four columns show the rendered image $\mathcal{R}_c$, indicated bounding boxes, masked image $\mathcal{R}_m$, and completed image $\mathcal{R}_c'$. (b) Completed mesh models; the top and bottom rows for each region show the original models $\mathcal{M}$ and completed models $\mathcal{M}'$, respectively.}
	\label{fig:results_swjtu}
\end{figure}

\begin{figure}[H]
	\centering
	\subcaptionbox{Image completion}[0.26\linewidth]{
		\includegraphics[width=\linewidth]{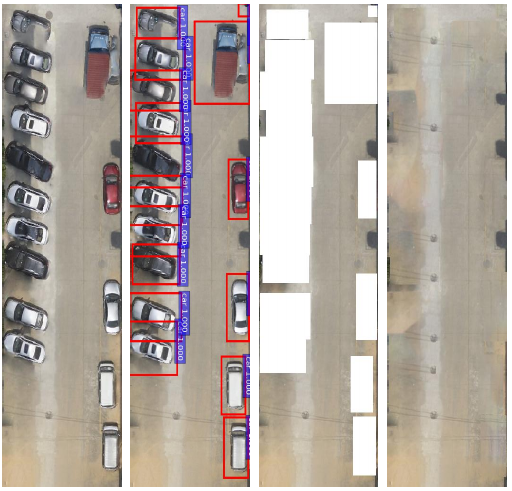}\vspace{0.2em}
		\includegraphics[width=\linewidth]{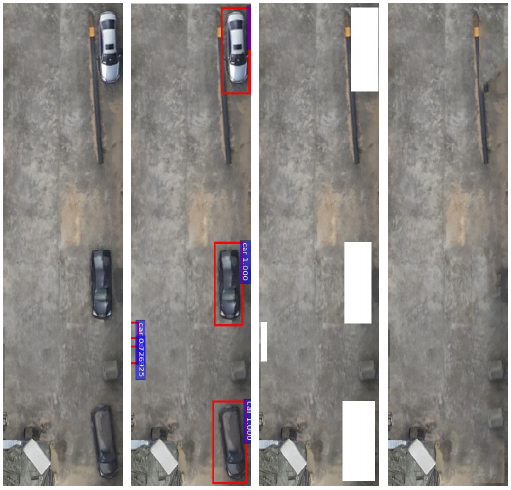}
	}
	\subcaptionbox{Mesh completion}[0.5\linewidth]{
		\includegraphics[width=\linewidth]{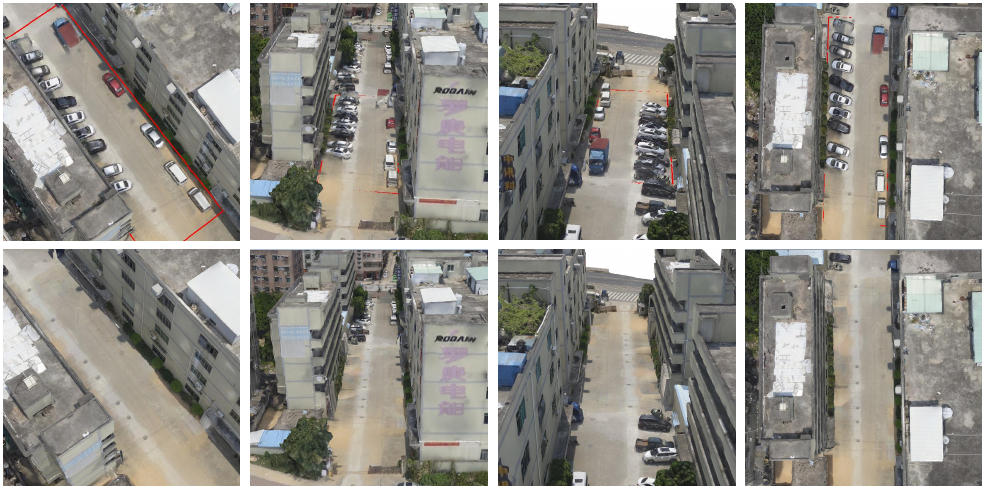}\vspace{0.2em}
		\includegraphics[width=\linewidth]{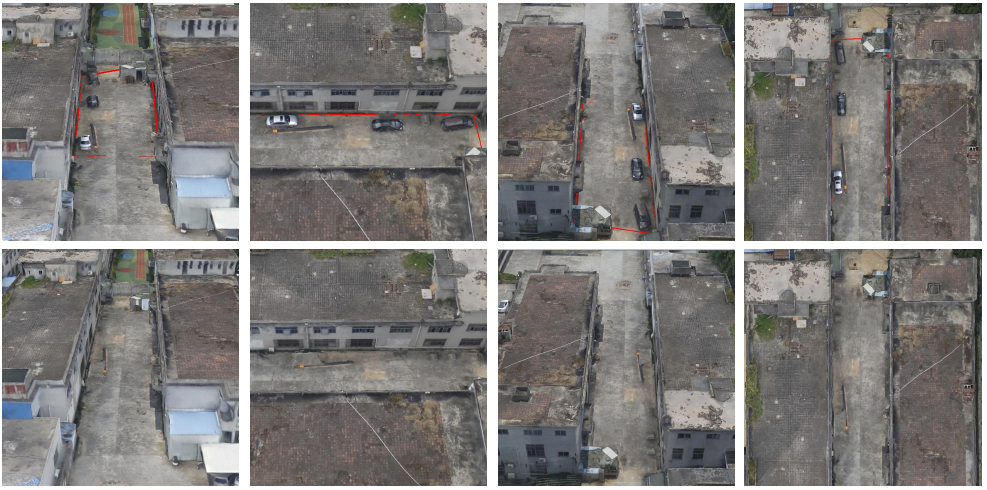}
	}
	\caption{Completed mesh models for the two regions of the Shenzhen dataset. (a) Completion of images; the four columns show the rendered image $\mathcal{R}_c$, indicated bounding boxes, masked image $\mathcal{R}_m$, and completed image $\mathcal{R}_c'$. (b) Completed mesh models; the top and bottom rows for each region show the original models $\mathcal{M}$ and completed models $\mathcal{M}'$, respectively.}
	\label{fig:results_shenzhen}
\end{figure}

\begin{figure}[H]
	\centering
	\subcaptionbox{Image completion}[0.26\linewidth]{
		\includegraphics[width=\linewidth]{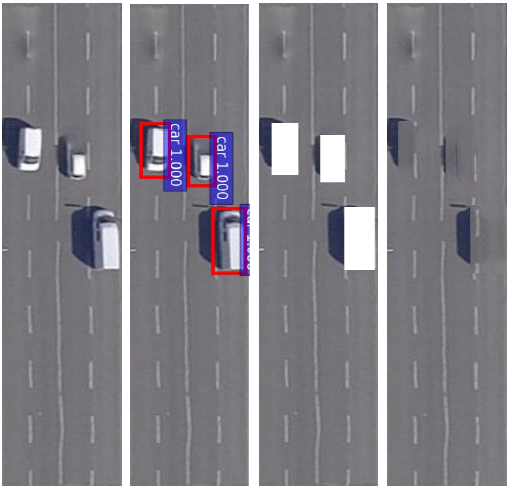}\vspace{0.2em}
		\includegraphics[width=\linewidth]{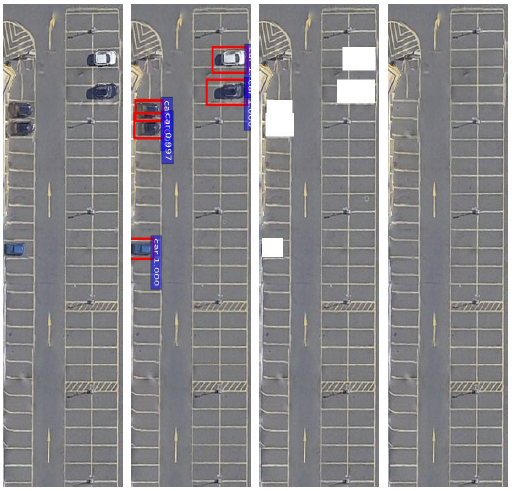}
	}
	\subcaptionbox{Mesh completion}[0.5\linewidth]{
		\includegraphics[width=\linewidth]{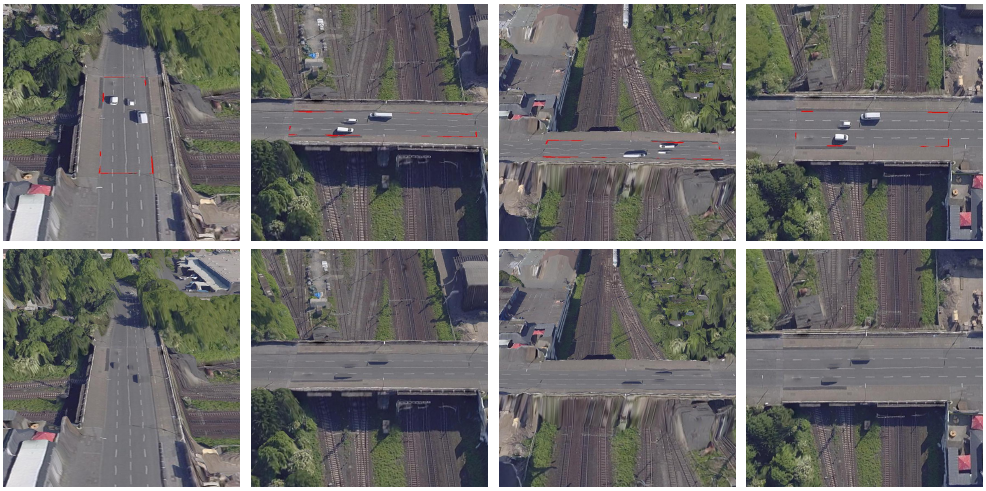}\vspace{0.2em}
		\includegraphics[width=\linewidth]{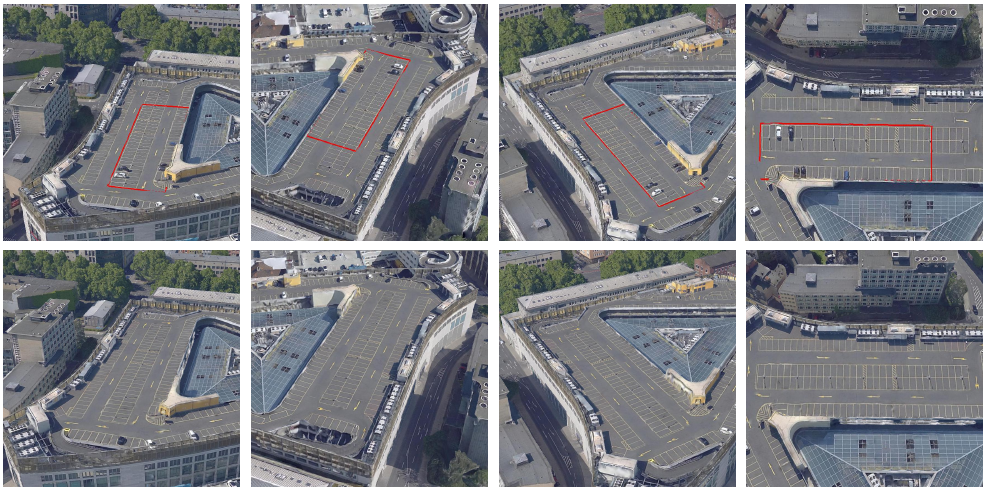}
	}
	\caption{Completed mesh models for the two regions of the Dortmund dataset. (a) Completion of images; the four columns show the rendered image $\mathcal{R}_c$, indicated bounding boxes, masked image $\mathcal{R}_m$, and completed image $\mathcal{R}_c'$. (b) Completed mesh models; the top and bottom rows for each region show the original models $\mathcal{M}$ and completed models $\mathcal{M}'$, respectively.}
	\label{fig:results_dort}
\end{figure}

Figures \ref{fig:results_swjtu} to \ref{fig:results_dort} show the completed results for both the images and mesh models.
Two regions for each dataset are separated into the top and bottom halves.
Except for the shadow regions in the Dortmund dataset, the results are quite satisfactory.
The road markers are recovered reasonably well, such as the crosswalk and X-cross in the SWJTU dataset, and the structured labels in the parking lot of the Dortmund dataset.
For the Shenzhen dataset, the faint structure patterns on the textureless road are also preserved.
The right part of each figure shows the mesh models before and after completion from four different viewpoints.
The proposed methods can produce cleaner road scenes after flattening the geometries and completing the textures of the mesh models. More supplementary video demonstrations involving large regions are available at \url{https://vrlab.org.cn/~hanhu/projects/mesh}.

\subsection{Comparison of image completion}
\subsubsection{Qualitative comparison}

\begin{figure}[H]
	\centering
	\subcaptionbox{Input}[0.19\linewidth]{
		\includegraphics[width=\linewidth]{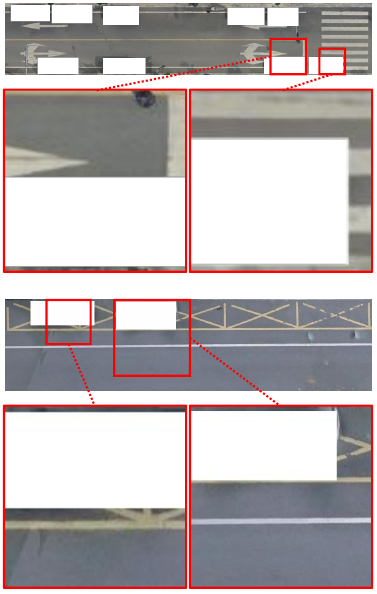}
	}
	\subcaptionbox{Reference}[0.19\linewidth]{
		\includegraphics[width=\linewidth]{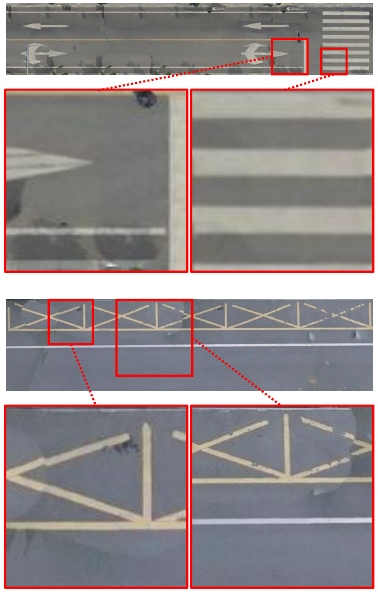}
	}
	\subcaptionbox{\cite{he2012statistics}}[0.19\linewidth]{
		\includegraphics[width=\linewidth]{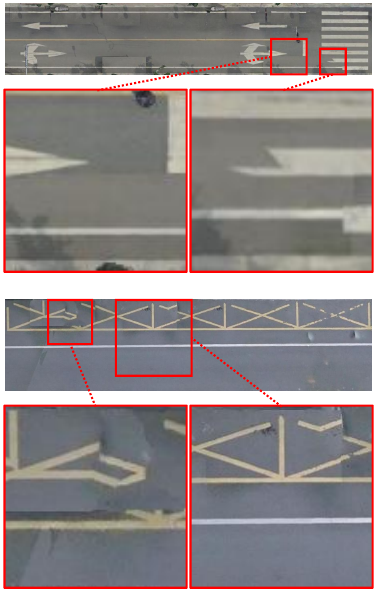}
	}
	\subcaptionbox{\cite{huang2014image}}[0.19\linewidth]{
		\includegraphics[width=\linewidth]{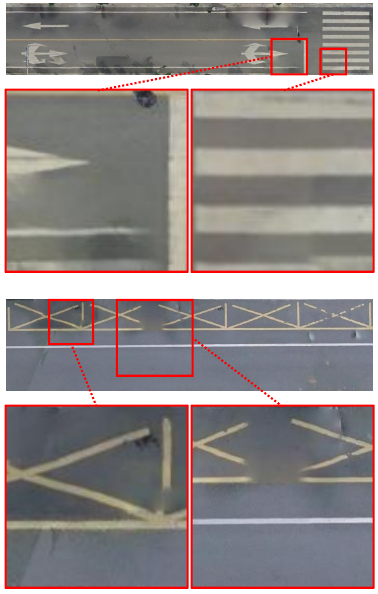}
	}
	\subcaptionbox{Proposed}[0.19\linewidth]{
		\includegraphics[width=\linewidth]{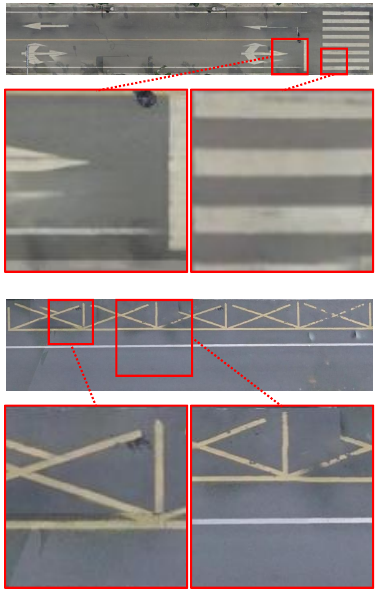}
	}
	
	\caption{Comparisons of different completion methods on the SWJTU dataset. The red rectangles indicate enlarged regions.}
	\label{fig:comparisons_swjtu}
\end{figure}

\begin{figure}[H]
	\centering
	\subcaptionbox{Input}[0.19\linewidth]{
		\includegraphics[width=\linewidth]{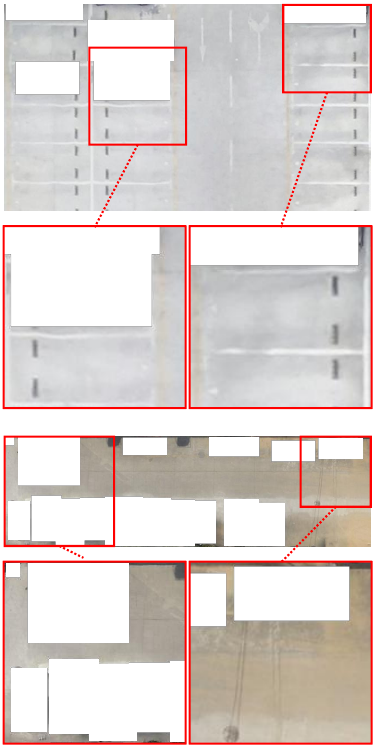}
	}
	\subcaptionbox{Reference}[0.19\linewidth]{
		\includegraphics[width=\linewidth]{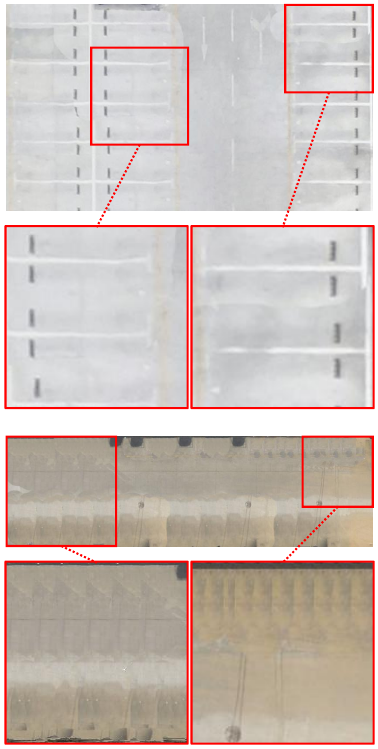}
	}
	\subcaptionbox{\cite{he2012statistics}}[0.19\linewidth]{
		\includegraphics[width=\linewidth]{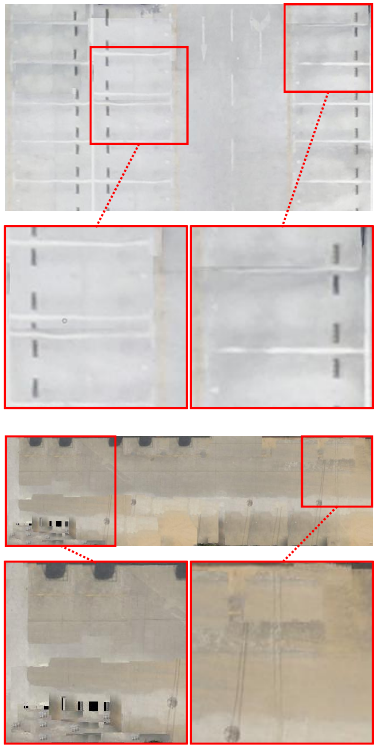}
	}
	\subcaptionbox{\cite{huang2014image}}[0.19\linewidth]{
		\includegraphics[width=\linewidth]{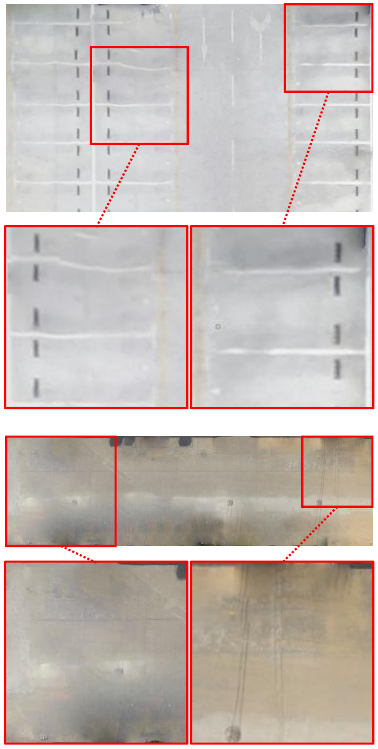}
	}
	\subcaptionbox{Proposed}[0.19\linewidth]{
		\includegraphics[width=\linewidth]{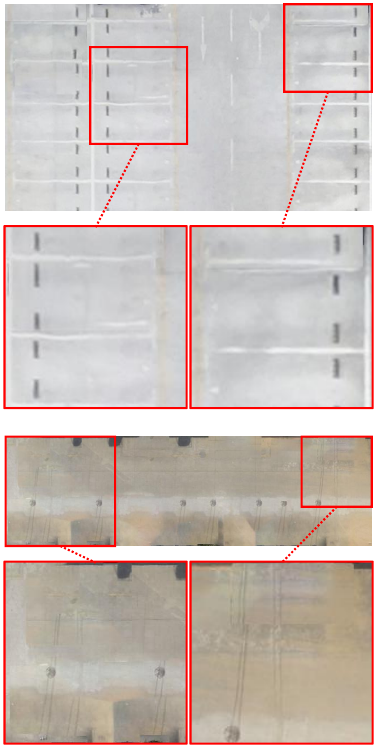}
	}
	
	\caption{Comparisons of different completion methods on the SWJTU dataset. The red rectangles indicate enlarged regions.}
	\label{fig:comparisons_shenzhen}
\end{figure}

\begin{figure}[H]
	\centering
	\subcaptionbox{Input}[0.19\linewidth]{
		\includegraphics[width=\linewidth]{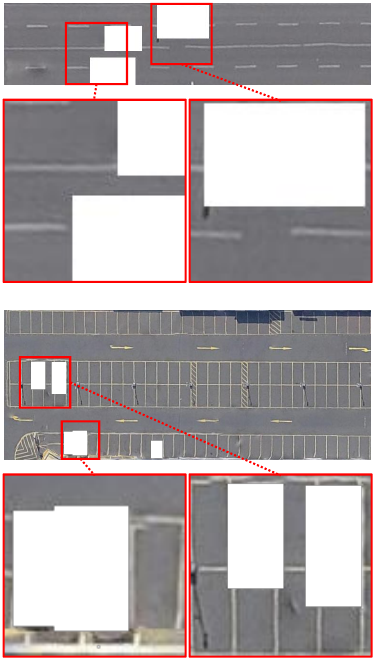}
	}
	\subcaptionbox{Reference}[0.19\linewidth]{
		\includegraphics[width=\linewidth]{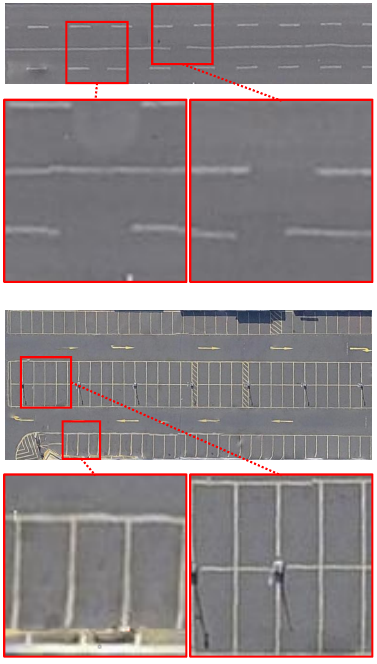}
	}
	\subcaptionbox{\cite{he2012statistics}}[0.19\linewidth]{
		\includegraphics[width=\linewidth]{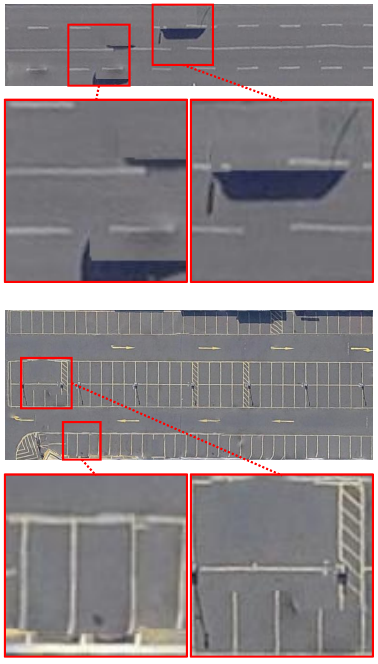}
	}
	\subcaptionbox{\cite{huang2014image}}[0.19\linewidth]{
		\includegraphics[width=\linewidth]{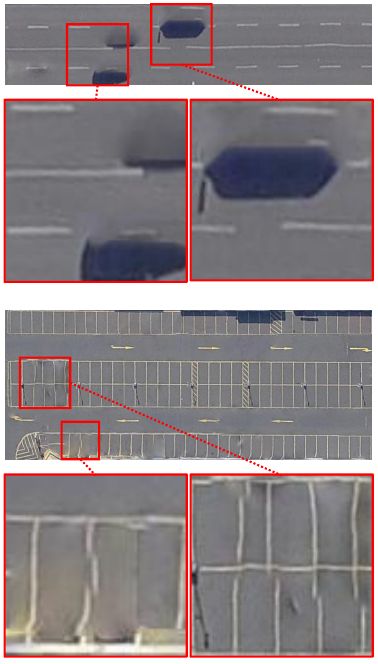}
	}
	\subcaptionbox{Proposed}[0.19\linewidth]{
		\includegraphics[width=\linewidth]{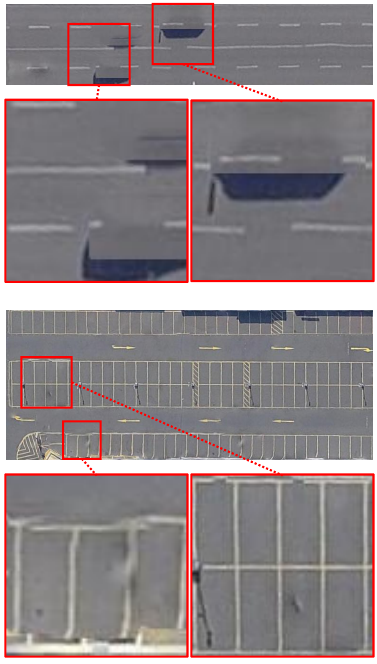}
	}
	
	\caption{Comparisons of different completion methods on the SWJTU dataset. The red rectangles indicate enlarged regions.}
	\label{fig:comparisons_dort}
\end{figure}

To evaluate the performance of image completion against state-of-the-art approaches, two approaches are considered: methods based on statistics of patch offsets \citep{he2012statistics}, and planar structure guidance \citep{huang2014image}. Figures \ref{fig:comparisons_swjtu} to \ref{fig:comparisons_dort} show the image completion results of the algorithms mentioned above. The first column shows images to be completed; the second column shows reference results edited manually using Adobe Photoshop \citep{adobe2020photoshop}. Zoomed regions are also shown to provide details of the completion results.

By comparing the results on different datasets, it can be seen that the results by \cite{he2012statistics} exhibit artificial noise, particularly the result under the simplest scenario in Shenzhen. In addition, although the method in \cite{huang2014image} yields better results than that in \cite{he2012statistics}, blurry areas occur, and perfect linear features cannot be preserved in some experiments. The proposed method exhibits better performance than the other two methods. Specifically, distinct linear features are preserved to achieve desirable completion results, as for example, the zebra crossing and parking lines.

\subsubsection{Quantitative comparison}

The peak signal-to-noise ratio (PSNR) is the most common and widely used objective evaluation index for images. Structural similarity (SSIM) is another image quality evaluation index. It measures image similarity in terms of brightness, contrast, and structure. Herein, we evaluate the completion results on road area texture integration images in terms of PSNR and SSIM, as in \citep{hore2010image}.
We consider a manually repaired image using Adobe Photoshop as a reference to calculate the PSNR and SSIM of different methods. Table \ref{tab:completion_evaluation} shows these evaluation indexes \citep{hore2010image}. It can be seen that the proposed method has better average PSNR and SSIM than the other two state-of-the-art methods. Regarding PSNR, the proposed method achieved the best results for all the testing samples; regarding SSIM, it achieved the best results for four out of six samples.

\begin{table}[h]
    \centering
    \caption{Comparison with other approaches in terms of PSNR and SSIM on the six datasets. The best results are highlighted in bold.}
    \label{tab:completion_evaluation}
    \resizebox{\textwidth}{!}{
    \begin{tabular}{c|c|cccccc|c}
      \hline
      \multicolumn{2}{c|}{Dataset}                        & SWJTU1 & SWJTU2 & Shenzhen1 & Shenzhen2 & Dortmund1 & Dortmund2 & Average \\ \hline
      \multirow{3}{*}{PSNR} & Proposed method             &  \textbf{27.49}  & \textbf{25.40}  & \textbf{26.56}     & \textbf{24.98}     & \textbf{29.67}     & \textbf{33.88}     & \textbf{28.00}   \\
                            & \cite{he2012statistics} & 26.92  & 25.30  & 26.20     & 23.16     & 29.45     & 32.94     & 27.33   \\
                            & \cite{huang2014image}   & 25.38  & 24.51  & 22.54     & 22.84     & 27.09     & 32.42     & 25.80   \\ \hline
      \multirow{3}{*}{SSIM} & Proposed method             & \textbf{0.932}  & \textbf{0.888}  & 0.918     & \textbf{0.894}     & 0.950     & \textbf{0.975}     & \textbf{0.926}   \\
                            & \cite{he2012statistics} & 0.919  & 0.878  & \textbf{0.926}     & 0.883     & \textbf{0.954}     & 0.969     & 0.921   \\
                            & \cite{huang2014image}   & 0.925  & 0.882  & 0.920     & 0.866     & 0.947     & 0.954     & 0.916   \\ \hline
      \end{tabular}
    }
\end{table}

\subsection{Analysis of linear guidance and directional constraint}

\begin{figure}[H]
	\centering
	\subcaptionbox{Proposed method}[0.24\linewidth]{
		\includegraphics[width=\linewidth]{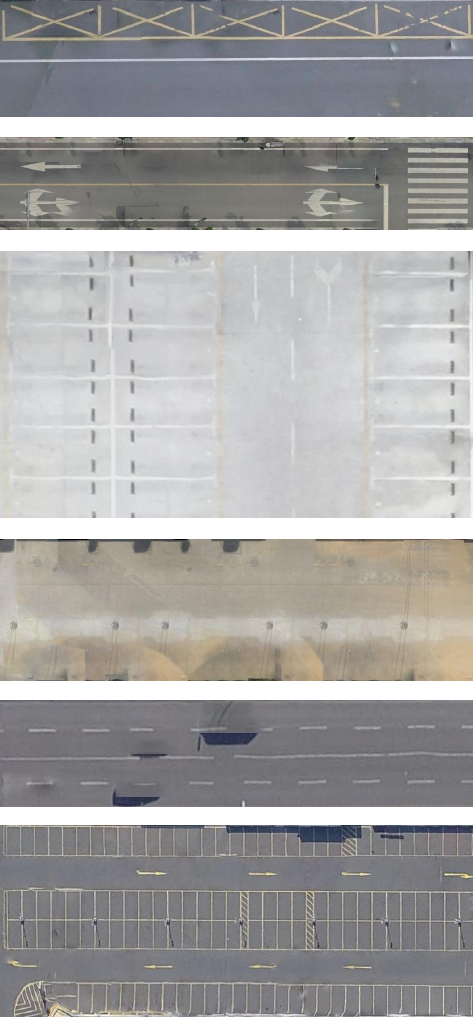}
	}
	\subcaptionbox{w/o. Directional guidance}[0.24\linewidth]{
		\includegraphics[width=\linewidth]{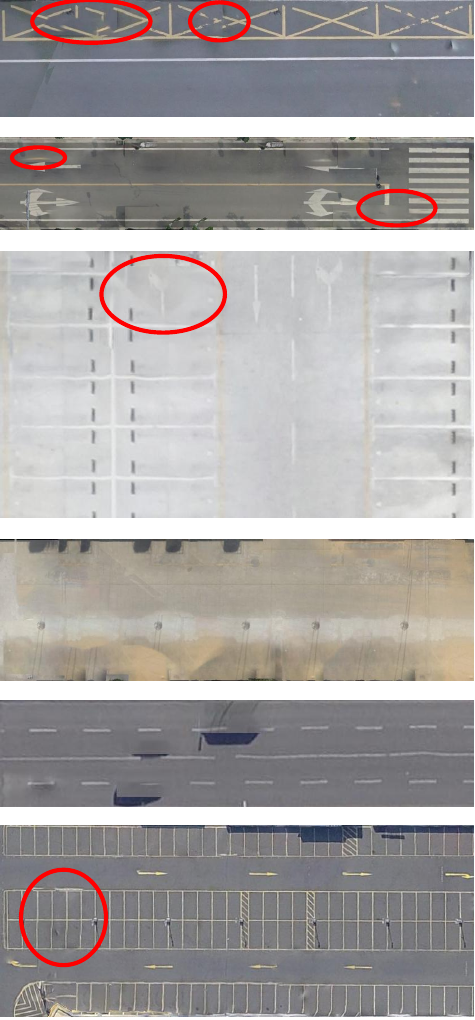}
	}
	\subcaptionbox{w/o. Linear ordering}[0.24\linewidth]{
		\includegraphics[width=\linewidth]{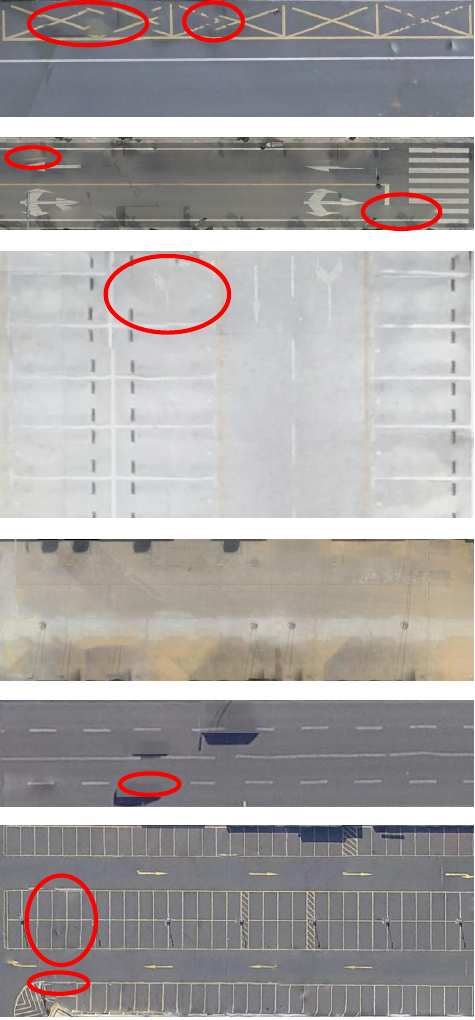}
	}
	\subcaptionbox{w/o. Both}[0.24\linewidth]{
		\includegraphics[width=\linewidth]{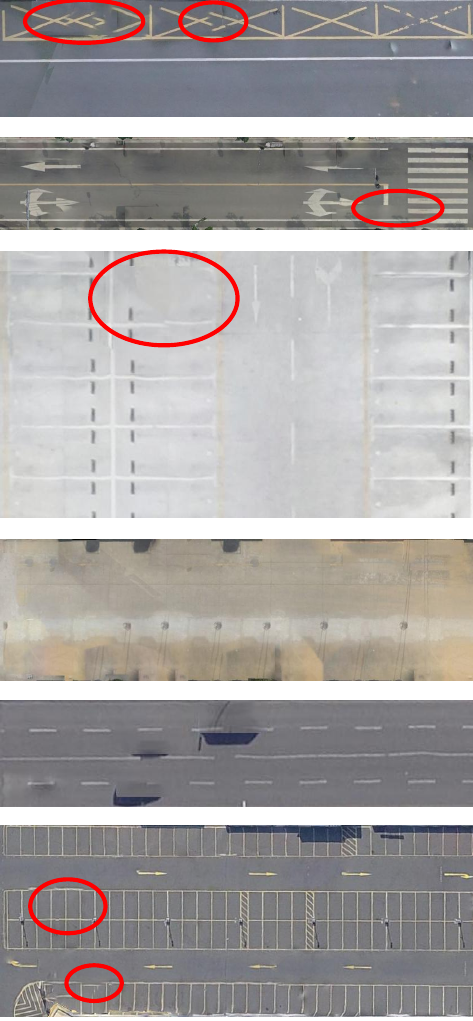}
	}
	\caption{Completion results under different settings. Highlighted regions are where the errors appear.}
	\label{fig:if_buffer_sort}
\end{figure}

The proposed method prioritizes the target patches and constrains the random searching area to improve the completion result.
To evaluate the effect of the priority setting by linear guidance and directionally constrained searching, we conducted ablation studies. Figure \ref{fig:if_buffer_sort} shows typical comparisons between methods under different settings.

\begin{table}[H]
	\centering
	\caption{Completion results evaluation under different algorithm settings. The best results are highlighted in bold.}
	\label{tab:evaluation_indexes}
	\resizebox{\textwidth}{!}{
		\begin{tabular}{c|c|cccccc}
			\hline
			Evaluation            & Method                    & SWJTU1           & SWJTU2           & Shenzhen1        & Shenzhen2        & Dortmund1       & Dortmund2        \\ \hline
			\multirow{4}{*}{PSNR} & Proposed                  & \textbf{27.27} & \textbf{25.67} & \textbf{25.24} & \textbf{28.96} & 28.36 & \textbf{34.11} \\
			& w/o. Directional Guidance & 26.41          & 24.67          & 25.03          & 28.77          & \textbf{28.49}         & 32.28          \\
			& w/o. Linear Ordering      & 26.53          & 24.86          & 25.10          & 28.55          & 28.48          & 32.33          \\
			& w/o. Both                 & 26.15          & 25.00          & 24.94          & 28.45          & 28.48          & 34.07          \\ \hline
			\multirow{4}{*}{SSIM} & Proposed                  & \textbf{0.930}  & \textbf{0.893}  & 0.901  & \textbf{0.944}           & 0.935           & \textbf{0.976}  \\
			& w/o. Directional Guidance & 0.912           & 0.880           & 0.899           & \textbf{0.944}  & 0.945       & 0.966           \\
			& w/o. Linear Ordering      & 0.915           & 0.876           & \textbf{0.902}           & 0.941           & 0.945          & 0.966           \\
			& w/o. Both                 & 0.910           & 0.878           & 0.900           & 0.942           & \textbf{0.947}        & 0.975           \\ \hline
		\end{tabular}
	}
\end{table}

By comparing the results in Figure \ref{fig:if_buffer_sort}, it can be concluded that the two strategies remarkably improve the completion results in complex scenarios with numerous linear structures. In addition, these methods yield similar results in a simple scenario, for example, Shenzhen1. From Table \ref{tab:evaluation_indexes}, it can be seen, that except for Shenzhen1 and Dortmund1, the proposed method yields the best completion results. The failure of the proposed method in these datasets can be attributed to the fact that both datasets have few artificial linear structures. In fact, it is also demonstrated that the guidance of a linear feature is valid in image completion of urban road areas, which always have numerous linear features.

\subsection{Discussion and limitations}

It can intuitively be seen that the proposed mesh completion method yields superior results, which are difficult to obtain by using UV editing tools in modeling software. After completion, regions with no useful texture image, for example, regions occluded by cars, can obtain a plausible texture. The quantitative experiments demonstrate that synthetic texture is sufficient as a model exhibition. Moreover, the proposed structure-aware image completion algorithm can achieve effects similar to those by traditional manual repair methods using Photoshop, with little interaction and higher automation. It is also more effective than state-of-the-art methods \citep{he2012statistics, huang2014image} in image completion of urban roads.

\red{With regard to the time complexities, for steps of model loading and updating (from disks), vehicle detection and texture integration, they could give almost near real-time feedbacks. For step of image completion, currently it is implemented in Matlab based on existing implementation \citep{huang2014image} and costs about one minute for a patch in the experiments. However, Patch Match has linear time complexity with regard to the number of pixels \citep{barnes2009patchmatch}, and the running time could be significantly accelerated using implementation by native programming languages, such as C/C++.}

Despite the good completion results, there are certain limitations. First, the proposed image completion algorithm is not always stable because of the random search. However, we improved it compared with state-of-the-art methods \citep{barnes2009patchmatch, huang2014image}. Second, when blank regions have no connection with the surrounding pixels, the proposed completion algorithm cannot be applied. In the future, we will conduct related research on texture generation in semantic models.

\section{Conclusion}
\label{s:conclusion}

\red{The photogrammetric mesh models, especially produced from aerial oblique images, have become probably the most important approach for city-scale modeling \citep{google2020earth}.
However, the problems of noises and defects of photogrammetric mesh models also raise a large amount of attentions in the community.
This paper deals with the undesired objects in the geometries and textures of the models, targeting the urban road scenarios.
Specifically, we propose a practical strategy to handle the discontinuous texture atlases by directly rendering the mesh models using graphics pipeline.
An improved PatchMatch-based image completion approach is also proposed to fill the textures of vehicles.
In addition, the proposed strategy can also handle tiled models, which are severely fragmented.
Future directions on the processing of photogrammetric mesh models may include: (1) adoption of similar strategy for the editing of other objects, such as distorted fa\c{c}ades; (2) other general purpose approaches for the simplification and regularization of the mesh models \citep{li2021featurepreserving}; and (3) enriching the mesh models with semantic information \citep{zhu2017robust,rouhani2017semantic}.}

\section*{Acknowledgments}
This work was supported in part by the National Key Research and Development Program of China (Project No. 2018YFC0825803) and by the National Natural Science Foundation of China (Project No. 41631174, 42071355, 41871291). In addition, the authors gratefully acknowledge the provision of the datasets by ISPRS and EuroSDR, which were released in conjunction with the ISPRS Scientific Initiatives 2014 and 2015, led by ISPRS ICWG I/Vb.

\bibliographystyle{model2-names}
\bibliography{TextureEditing}

\end{document}